\documentclass[runningheads]{llncs}
\usepackage[T1]{fontenc}
\usepackage{graphicx}
\usepackage[numbers,sort&compress]{natbib}
\usepackage{todonotes}

\usepackage{makecell}
\usepackage{xcolor,pifont}
\usepackage{bbding}
\usepackage{multirow}
\usepackage[caption=false]{subfig}
\newcommand*\colourcheck[1]{%
  \expandafter\newcommand\csname #1check\endcsname{\textcolor{#1}{\ding{52}}}%
}
\colourcheck{green}

\begin{document}
\title{FinnWoodlands Dataset}
%
%
\author{Juan Lagos\inst{1} \and
Urho Lempiö\inst{1} \and
Esa Rahtu\inst{1}\orcidID{0000-0001-8767-0864}}


%
\authorrunning{J. Lagos et al.}
%
\institute{Tampere University, Tampere, Finland \\
\email{\{juanpablo.lagosbenitez, urho.lempio, esa.rahtu\}@tuni.fi}}


%
\maketitle              
\begin{abstract}

While the availability of large and diverse datasets has contributed to significant breakthroughs in autonomous driving and indoor applications, forestry applications are still lagging behind and new forest datasets would most certainly contribute to achieving significant progress in the development of data-driven methods for forest-like scenarios.  This paper introduces a forest dataset called \textit{FinnWoodlands}, which consists of RGB stereo images, point clouds, and sparse depth maps, as well as ground truth manual annotations for semantic, instance, and panoptic segmentation. \textit{FinnWoodlands} comprises a total of 4226 objects manually annotated, out of which 2562 objects (60.6\%) correspond to tree trunks classified into three different instance categories, namely "Spruce Tree", "Birch Tree", and "Pine Tree". Besides tree trunks, we also annotated "Obstacles" objects as instances as well as the semantic stuff classes "Lake", "Ground", and "Track". Our dataset can be used in forestry applications where a holistic representation of the environment is relevant. We provide an initial benchmark using three models for instance segmentation, panoptic segmentation, and depth completion, and illustrate the challenges that such unstructured scenarios introduce. \textit{FinnWoodlands} dataset is available at \url{https://github.com/juanb09111/FinnForest.git}.

\end{abstract}

\section{Introduction}

Public datasets have contributed  to attracting interest in research in the field of computer vision. Data availability has accelerated the development of new paradigms, techniques, and models, especially data-driven methods and deep learning models designed to solve different computer vision tasks such as image segmentation, object detection, depth estimation, flow estimation, and object tracking \cite{review}. Some public datasets have even become evaluation benchmarks, allowing different methods to claim state-of-the-art status  \cite{coco, cityscapes, kitti, inet, nyu, pascal_voc, vkitti, semantic_kitti, BDD100K, mapillary, d2city, camvid, nusc, a2d2}. While most popular datasets are collected from urban environments, with clear benefits for the development of autonomous driving applications, other contexts are lagging behind in terms of the availability of data, more specifically, off-road landscapes like forests. Moving through the forest with a proper data collection setup is all the more challenging as compared to urban scenarios, and that explains, to some extent, the gap between the availability of forest datasets compared to indoor or urban datasets.

Nonetheless, numerous applications would benefit from increasing the availability of forest datasets. For instance, in agricultural, farming, and exploration applications \cite{robots, agri, dig_farm, harv_rob}, objects are less structured and defined, boundaries between objects are less clear and the movement, interaction, and appearance are fuzzier as compared to urban and indoor scenarios. This makes tasks such as autonomous navigation in forest-like scenarios very challenging since it is a less controlled environment with no lane lines, navigation signs, or clear paths. 

Moreover, navigation in such unstructured scenarios is less standardized. The variety of scenarios ranges from dirt and gravel roads in the middle of the forest, where larger vehicles  and heavy machinery like harvesters and tractors used in agricultural applications \cite{phdthesis} can be driven, to smaller trekking trails for hikers and explorers only. While forest datasets are of high relevance within a specific group of applications, more specifically in the forestry industry, there are common features in such datasets e.g. structureless nature, that other applications might possibly exploit.

Collecting data from forests is not a trivial task. Forests change significantly during different seasons, and the type of vegetation varies depending on the geographic location. Hence, providing an optimal forest dataset most certainly requires a collective effort. 
Motivated by this, we introduce \textbf{\textit{FinnWoodlands}}, a dataset collected from trekking trails in the forests of Finland. It consists of $5170$ stereo RGB frames, the corresponding LIDAR point clouds, and sparse depth maps for each frame. Besides, we provide semantic segmentation, instance segmentation, and panoptic segmentation annotations for $300$ frames that contain $4226$ objects annotated manually using CVAT \cite{cvat}, an open-source tool for image and video annotation.  We provide guidelines for other scientists to extend FinnWoodlands with more frames which would also increase variability in our dataset with forest images from different parts of the world. In that sense, FinnWoodlands sets up a robust seed dataset for forestry applications, with which we expect to attract the attention of the community of data scientists.

We evaluated our dataset with three different models for instance segmentation, panoptic segmentation and depth completion, namely Mask R-CNN \cite{maskrcnn}, EfficientPS \cite{effps}, and FuseNet \cite{2d3d}, respectively. We thus set an initial benchmark for our dataset. The major contributions of our work are the following:

\begin{itemize}
  \item We provide a forest dataset named \textit{FinnWoodlands} that consists of RGB stereo frames, point clouds, and sparse depth maps, as well as ground truth (GT) annotations for semantic segmentation, instance segmentation, and panoptic segmentation. To our best knowledge, no other dataset in the context of forestry applications provides panoptic segmentation GT annotations. Thus, we aim to facilitate the research in holistic scene understanding in forest environments. 
  \item We illustrate how to collect data in scenarios where mobility and navigation are challenging with a simple data collection setup that consists of a LIDAR sensor and a stereo camera  mounted on a backpack. Our setup can be easily replicated elsewhere to collect reliable data compared to other, more expensive solutions.
\end{itemize}

\section{Related Works}

\begin{figure}[!ht]
\centering
\begin{tabular}{ccc}
{ \thead{Dataset}} &
\thead{RGB Frame} &
\thead{Semantic Segmentation} \\

{{\rotatebox[origin=lB]{90}{RELLIS-3D \cite{rellis}}}}&
{\includegraphics[width = 1.7in, height = 0.8in]{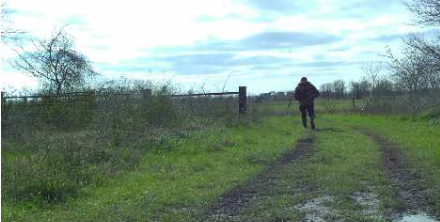}} &
{\includegraphics[width = 1.7in, height = 0.8in]{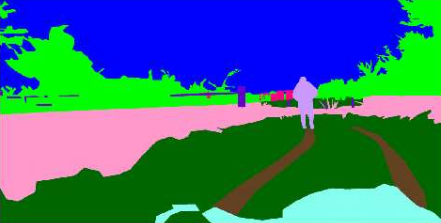}}\\

{{\rotatebox[origin=lB]{90}{RUGD \cite{rugd}}}}&
{\includegraphics[width = 1.7in, height = 0.8in]{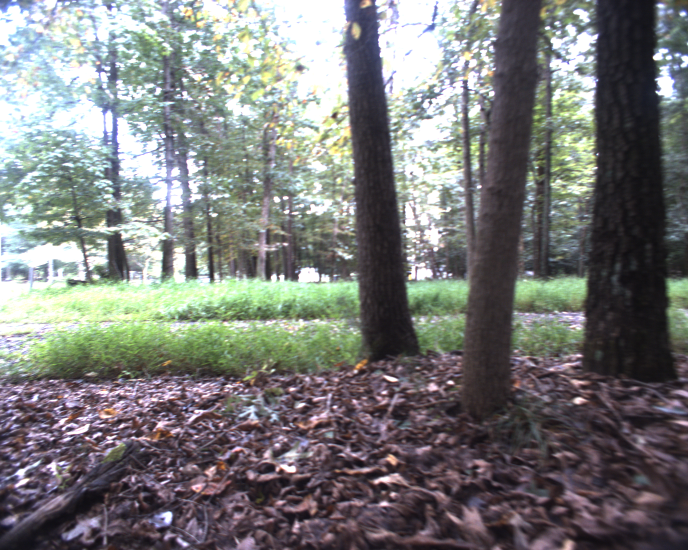}} &
{\includegraphics[width = 1.7in, height = 0.8in]{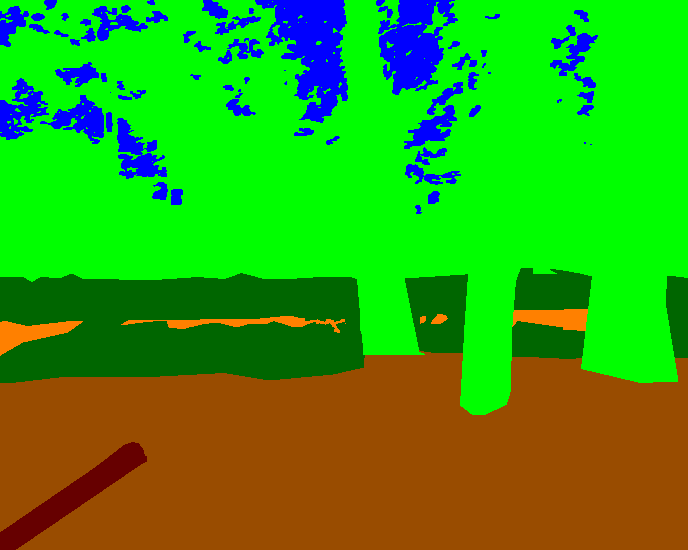}}\\

{{\rotatebox[origin=lB]{90}{TartanAir \cite{tartan}}}}&
{\includegraphics[width = 1.7in, height = 0.8in]{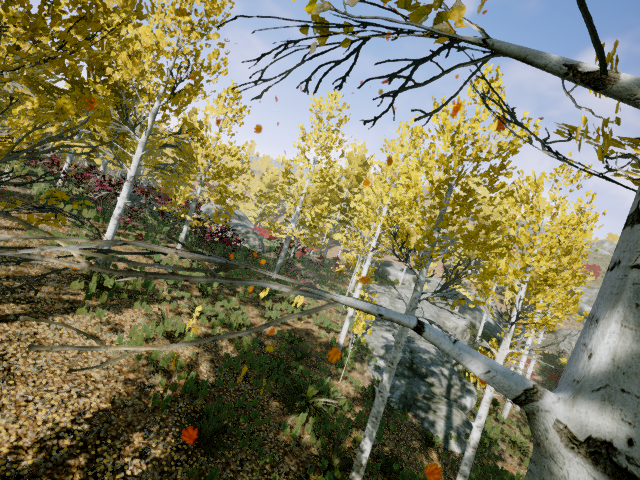}} &
{\includegraphics[width = 1.7in, height = 0.8in]{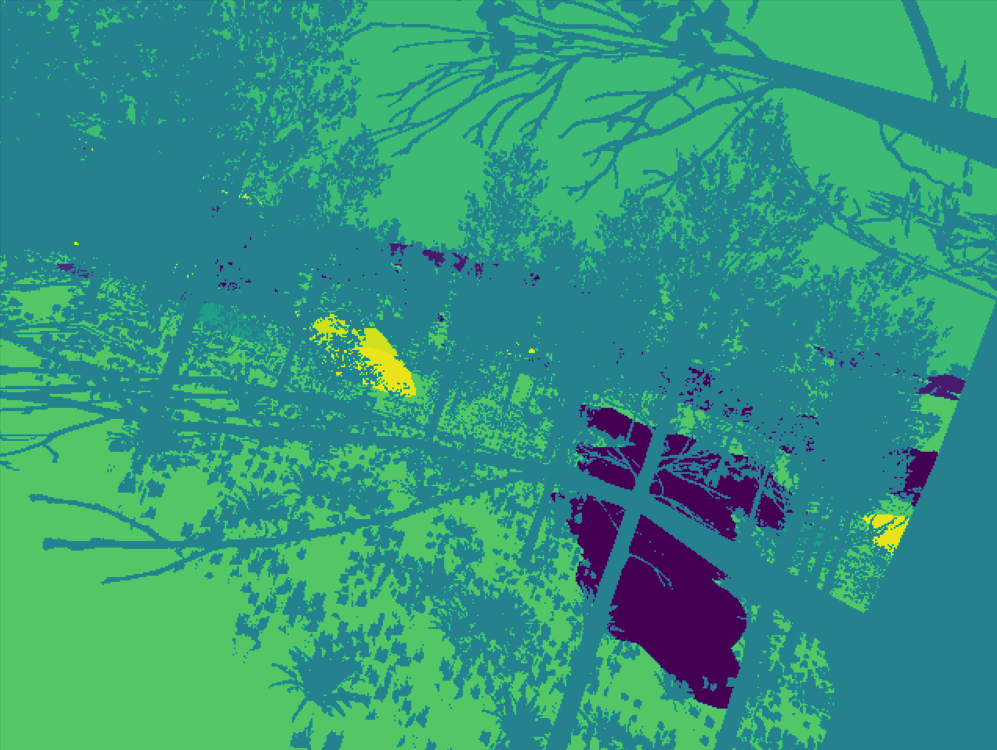}}\\

\end{tabular}
\caption{Semantic Segmentation Annotations. Sample RGB images and their corresponding semantic segmentation GT annotations from three forest datasets: RELLIS-3D \cite{rellis}, RUGD \cite{rugd}, andTartanAir \cite{tartan}.}
\label{datasets2}
\end{figure}

\begin{figure*}[!ht]
\centering
\begin{tabular}{ccc}
{ \thead{Dataset}} &
\thead{RGB Frame} &
\thead{Instance Segmentation \\ and Object Detection} \\

{{\rotatebox[origin=lB]{90}{CANATREE100 \cite{treedet}}}}&
{\includegraphics[width = 1.7in]{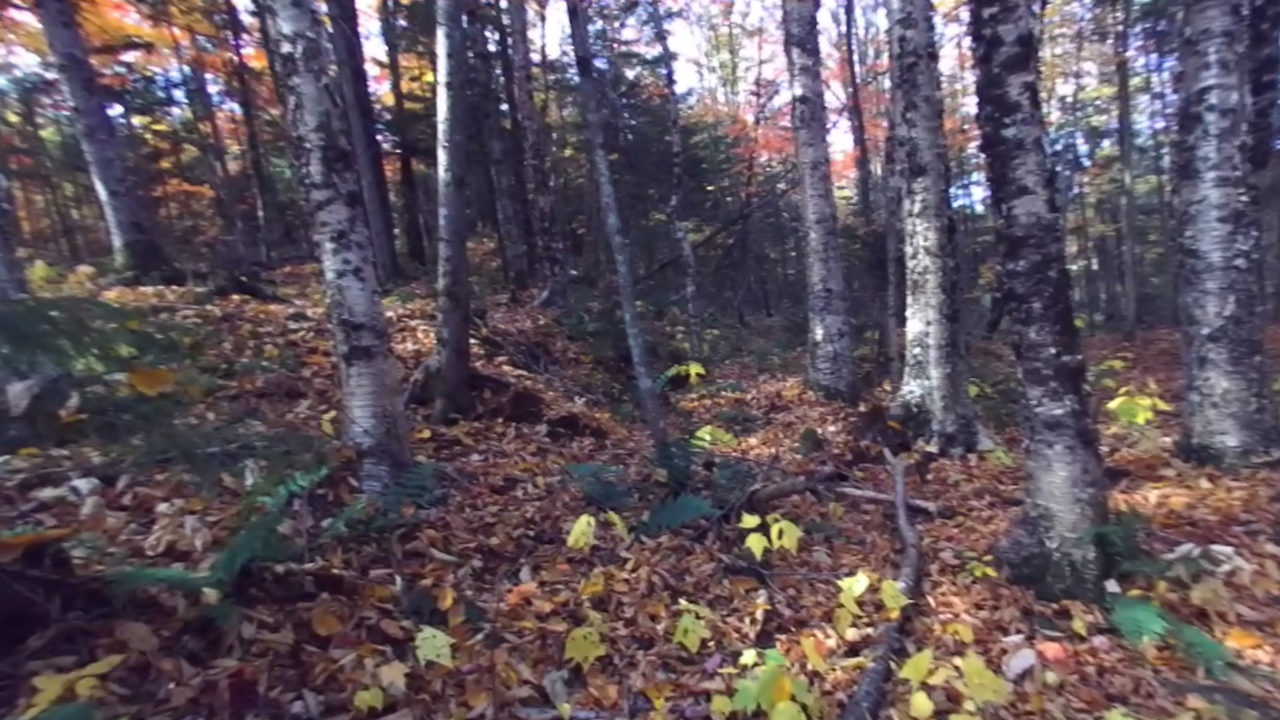}} &
{\includegraphics[width = 1.7in]{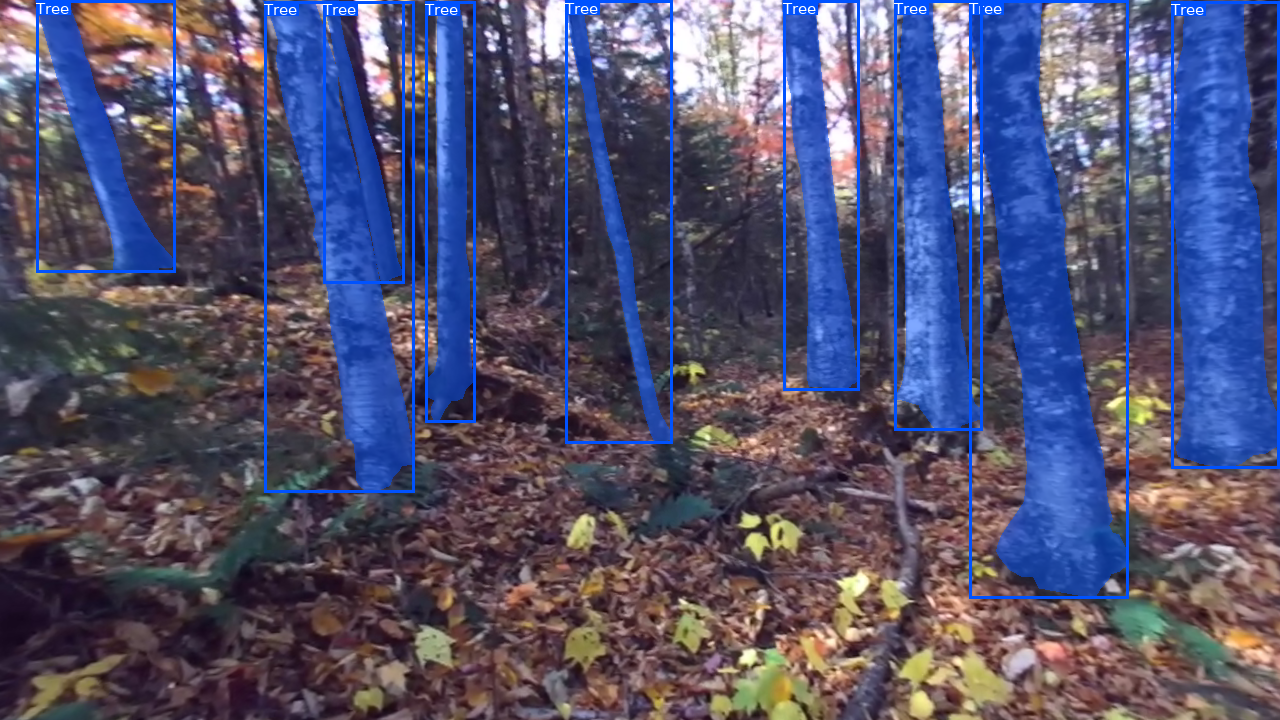}}\\

{{\rotatebox[origin=lB]{90}{ForTrunkDet \cite{fortrunk} }}}&
{\includegraphics[width = 1.7in]{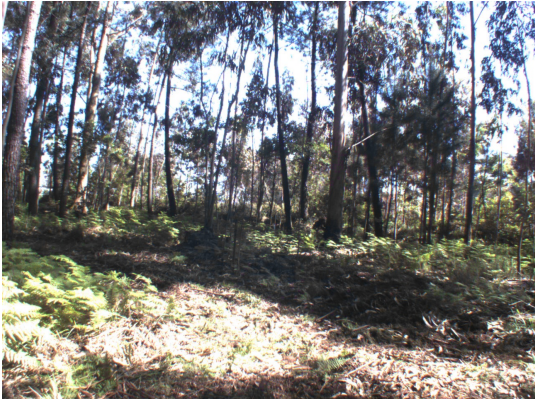}} &
{\includegraphics[width = 1.7in]{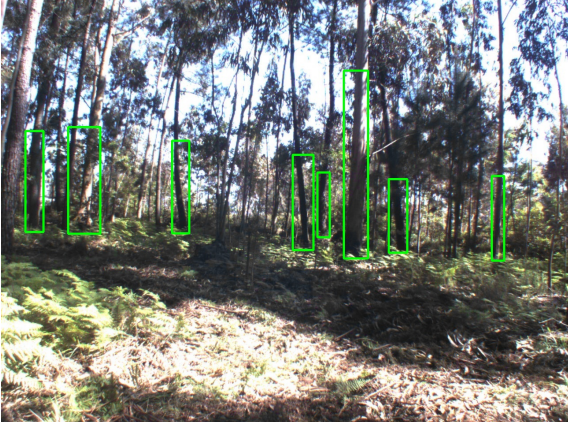}}\\

{{\rotatebox[origin=lB]{90}{SYNTHTREE43K \cite{treedet}}}}&
{\includegraphics[width = 1.7in, height=1.2in]{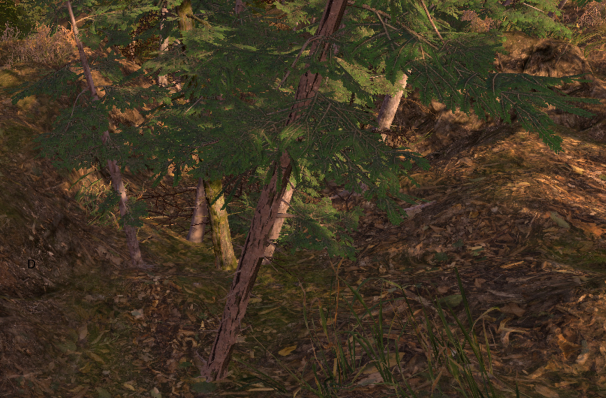}} &
{\includegraphics[width = 1.7in, height=1.2in]{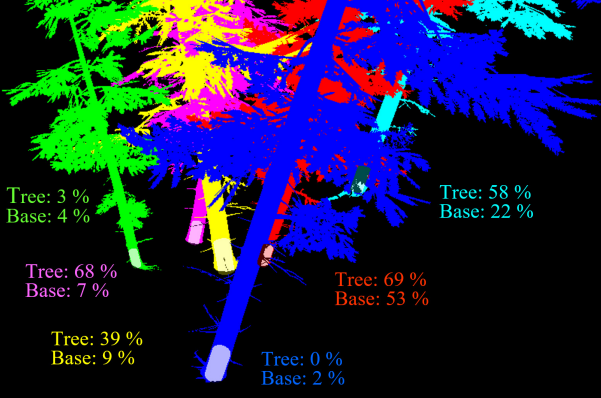}}\\

{{\rotatebox[origin=lB]{90}{TimberSeg \cite{timberseg}}}}&
{\includegraphics[width = 1.7in, height=1.05in]{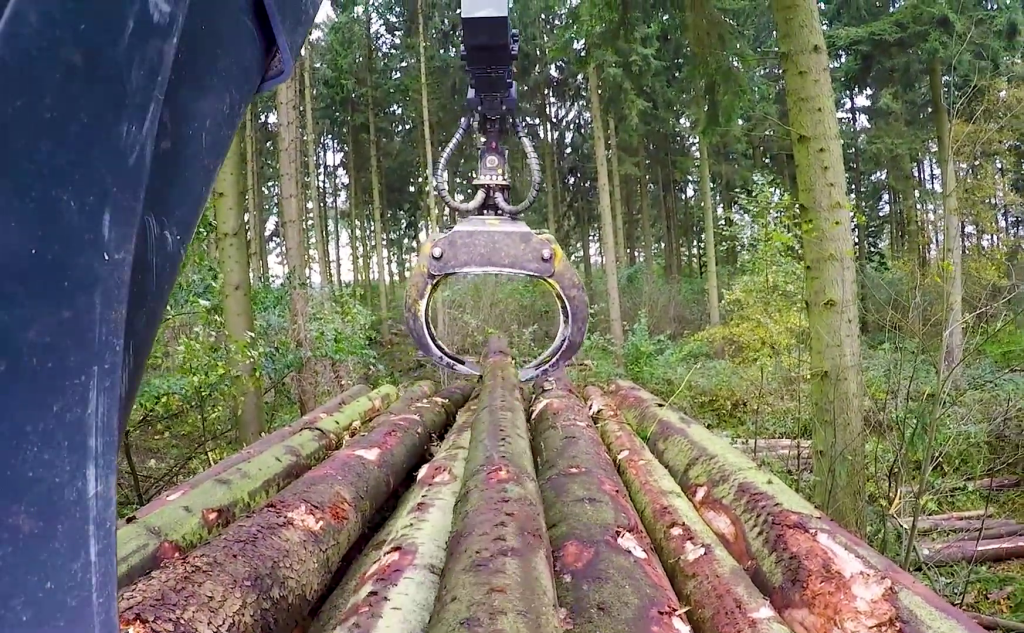}} &
{\includegraphics[width = 1.7in, height=1.05in]{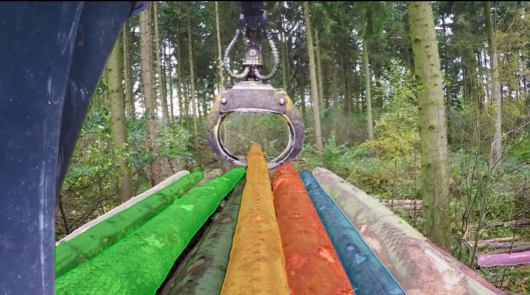}}\\

\end{tabular}
\caption{ Instance Segmentation Annotations. Sample RGB images and their corresponding instance segmentation GT annotations from four different forest datasets: CANATREE100 \cite{treedet}, ForTrunkDet \cite{fortrunk}, SYNTHTREE43K \cite{treedet}, TimberSeg \cite{timberseg}.}
\label{datasets}
\end{figure*}

Among numerous lists of public datasets, only a few are collected in forest scenarios, some of which are oriented to navigation tasks, while the rest focus on tree detection and segmentation for industrial forestry applications. Hereby, we present an overview of publicly available forest datasets and summary of the annotations provided by each one of the datasets shown in Table \ref{summaryanns}. Visualization of representative images and annotations for the listed datasets are depicted in Figure \ref{datasets2} and Figure \ref{datasets}.

\paragraph{\textbf{CANATREE100}} dataset \cite{treedet} consists of $100$ RGB and $100$ depth images and approximately 920 annotated trees. It was collected in the forests of Canada and provides annotations with segmentation masks for the tree trunks.

\paragraph{\textbf{ForTrunkDet}} \cite{fortrunk} is a dataset for tree trunk detection collected from three different forest locations. It consists of manually annotated visible and thermal images comprising two tree species corresponding to eucalyptus and pinus. It contains $2029$ images in the visible spectrum and $866$ thermal images.

\paragraph{\textbf{RELLIS-3D}} dataset \cite{rellis} consists of multi-modal synchronized sensor data frames collected from off-road environments. It is composed of five sequences that consist of RGB images, IMU data, GPS data, LIDAR point clouds, and stereo images. 

\paragraph{\textbf{The Robot Unstructured Ground Driving dataset (RUGD)}} \cite{rugd} is a dataset collected from a ground robot in semi-urban locations and unstructured scenarios like forests. It consists of video sequences containing objects with irregular and inconsistent geometric morphology. The robot moves around different types of terrain labeled as \textit{"creek"}, \textit{"park"}, \textit{"trail"}, and \textit{"village"}. 

\paragraph{\textbf{SYNTHTREE43K}} \cite{treedet} is a synthetic dataset collected from a simulated forest environment for tree detection with as many as $17$ different types of trees. It consists of $43k$ synthetic images produced using the Unity game engine \cite{unity}.

\paragraph{\textbf{TartanAir}} \cite{tartan} is a large photo-realistic dataset for navigation tasks. It provides outdoor and indoor scenes and various types of environments, including two subsets that contain forest scenes during different seasons, namely, autumn and winter. Due to its synthetic nature, TartanAir dataset also provides multimodal GT labels such as semantic segmentation tags, depth, camera pose, optical flow, stereo disparity, synthetic LIDAR points, and synthetic IMU readings.

\paragraph{\textbf{TimberSeg 1.0}} dataset \cite{timberseg} consists of 220 RGB images collected during different seasons and environment illumination conditions. It provides bounding box and segmentation mask annotations for individual tree logs. There are a total of $2500$ segmented logs in the \textit{TimberSeg 1.0} dataset. This dataset targets especially forwarding and log-picking applications.

\begin{table}
\centering
\caption{Datasets and GT Annotations}\label{summaryanns}
 \begin{tabular}{||c c c c c c c c ||} 
 \hline
 \thead{Dataset} &  \thead{{\rotatebox[origin=B]{78}{BBox}}} & {\rotatebox[origin=B]{78}{\thead{Instance \\ Segm}}}   & {\rotatebox[origin=B]{78}{\thead{Semantic \\ Segm}}} & {\rotatebox[origin=B]{78}{\thead{Panoptic \\ Segm}}} & {\rotatebox[origin=B]{78}{\thead{Point Cloud \\ Segm}}} & {\rotatebox[origin=B]{78}{\thead{Optical \\ Flow}}} & {\rotatebox[origin=B]{78}{\thead{Depth}}} \\ [0.5ex] 
 
 \hline\hline
CANATREE100 \cite{treedet} & \greencheck  & \greencheck &  \textcolor{red}{\XSolidBrush} &  \textcolor{red}{\XSolidBrush} & \textcolor{red}{\XSolidBrush} & \textcolor{red}{\XSolidBrush} & \greencheck \\
ForTrunkDet \cite{fortrunk} & \greencheck  &  \textcolor{red}{\XSolidBrush} & \textcolor{red}{\XSolidBrush} & \textcolor{red}{\XSolidBrush} & \textcolor{red}{\XSolidBrush} & \textcolor{red}{\XSolidBrush} & \textcolor{red}{\XSolidBrush}\\
RELLIS-3D \cite{rellis} & \textcolor{red}{\XSolidBrush} & \textcolor{red}{\XSolidBrush} & \greencheck & \textcolor{red}{\XSolidBrush} & \greencheck & \textcolor{red}{\XSolidBrush} & \textcolor{red}{\XSolidBrush}\\
RUGD \cite{rugd} &  \textcolor{red}{\XSolidBrush} & \textcolor{red}{\XSolidBrush} & \greencheck & \textcolor{red}{\XSolidBrush} & \textcolor{red}{\XSolidBrush} & \textcolor{red}{\XSolidBrush} & \textcolor{red}{\XSolidBrush}\\
SYNTHTREE43K \cite{treedet} & \greencheck & \greencheck & \textcolor{red}{\XSolidBrush} & \textcolor{red}{\XSolidBrush} & \textcolor{red}{\XSolidBrush} & \textcolor{red}{\XSolidBrush} & \greencheck\\ 
TartanAir \cite{tartan} &  \textcolor{red}{\XSolidBrush} & \textcolor{red}{\XSolidBrush} & \greencheck & \textcolor{red}{\XSolidBrush} & \greencheck & \greencheck & \greencheck\\ 
TimberSeg 1.0 \cite{timberseg} & \greencheck & \greencheck & \textcolor{red}{\XSolidBrush} & \textcolor{red}{\XSolidBrush} &\textcolor{red}{\XSolidBrush} & \textcolor{red}{\XSolidBrush} & \textcolor{red}{\XSolidBrush}\\ 
\textbf{FinnWoodlands (ours)} & \greencheck & \greencheck & \greencheck & \greencheck &\textcolor{red}{\XSolidBrush} & \textcolor{red}{\XSolidBrush} & \textcolor{red}{\XSolidBrush}\\[1ex] 
 \hline
 \end{tabular}
\end{table}

\section{Dataset Features}

\begin{figure}
\centering
\subfloat[Point Cloud]{%
  \includegraphics[clip,width= 0.9\columnwidth]{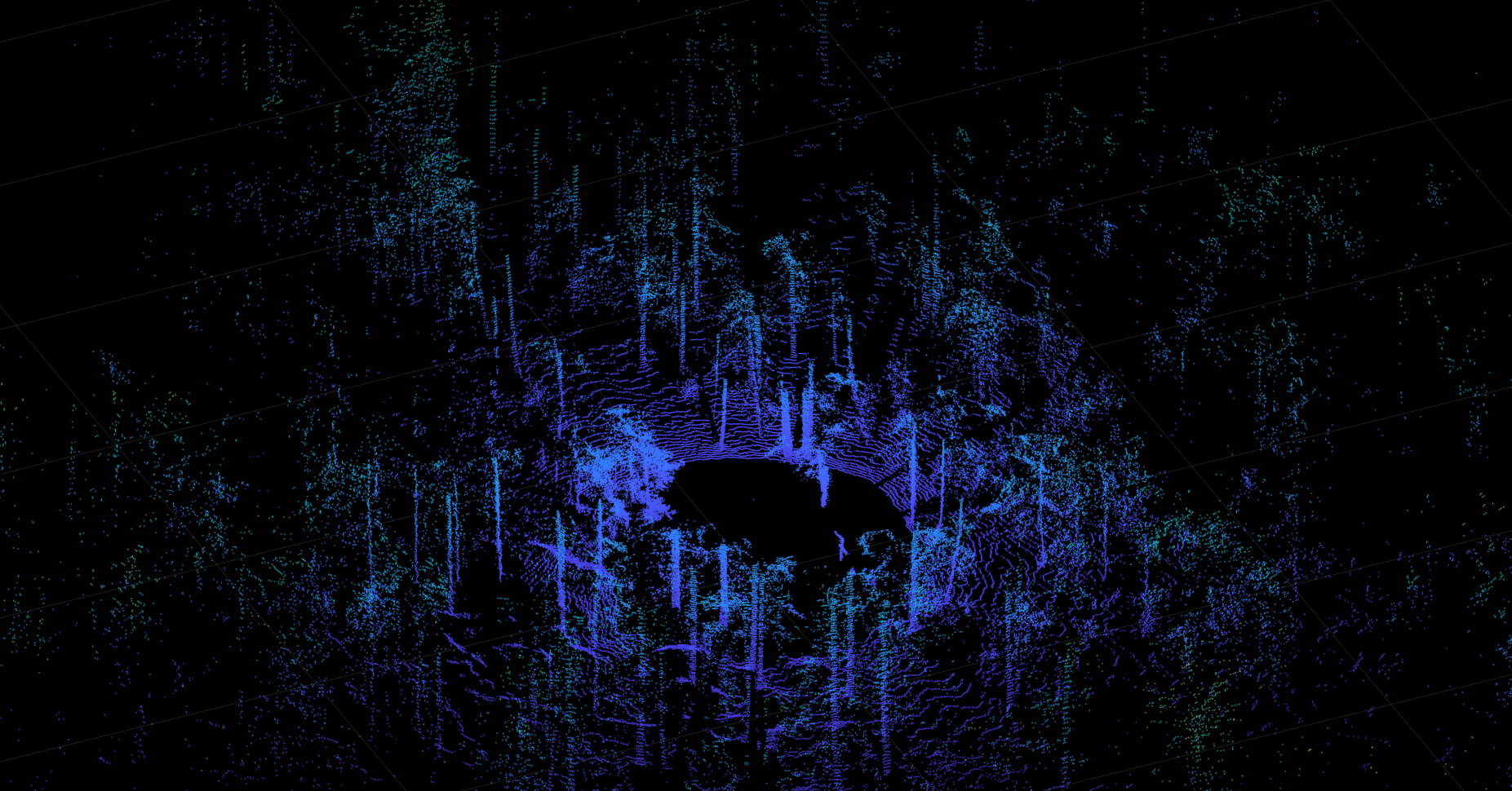}%
  \label{pc_proj_a}
}

\subfloat[Point Cloud Projection]{%
  \includegraphics[clip,width= 0.9\columnwidth]{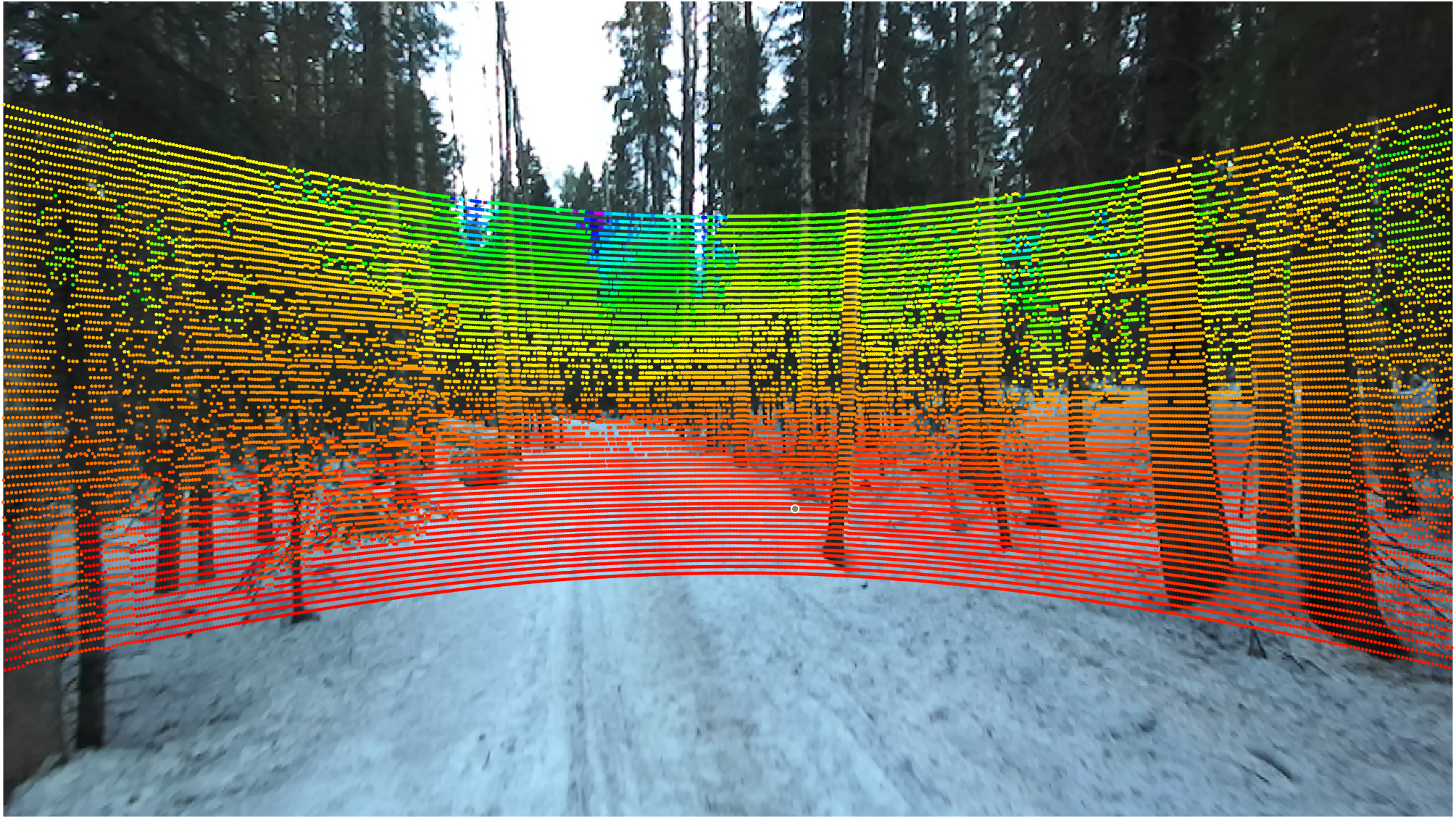}%
  \label{pc_proj_b}
}

\caption{\textit{FinnWoodlands} sample point cloud and sparse depth map visualization. The sparse depth map corresponds to the projection of the 3D point cloud onto to the 2D image.}
\label{pc_proj}
\end{figure}

Our dataset, \textit{FinnWoodlands}, comprises $5170$ synchronized stereo RGB images, LIDAR point clouds, and sparse depth maps for every frame. Figure \ref{pc_proj} shows a visualization of one sample frame with its corresponding point cloud (Figure \ref{pc_proj_a}), as well as the projection of the LIDAR points onto the image (Figure \ref{pc_proj_b}). The data was collected from three different locations near Tampere, Finland.  The annotations were done manually, consisting of semantic segmentation images, bounding boxes, class labels, instance segmentation, and panoptic segmentation GT images. We also provide COCO format annotation files for instance segmentation and panoptic segmentation. Table \ref{summary} shows a summary list of the data contained in our dataset.

\begin{table}[h!]
\centering
\caption{FinnWoodlands Data Summary}\label{summary}
 \begin{tabular}{||c c c ||} 
 \hline
 Data &  Number of samples & Format\\ [0.5ex] 
 \hline\hline
Stereo RGB Frames & 5170  & .jpg\\
LIDAR Point Clouds & 5170  & .pcd \\
Semantic GT & 300 & Label and RGB .png\\
Instance GT &  300 & RGB .png\\
Panoptic GT &  300 & RGB .png\\
Sparse Depth Maps & 5170 & Depth .png\\ [1ex] 
 \hline
 \end{tabular}
\end{table}

We used "stuff" and "things" objects as class categories in our dataset. In the context of computer vision, "stuff" classes refer to uncountable objects, for instance, "sky" or "ground", while "things" classes refer to countable objects such as "car" or "person" \cite{Adelson2001OnSS}. Objects under the category "things"  can be annotated with bounding boxes and instance segmentation masks, and objects under the category "stuff" are generally annotated with pixel-wise segmentation masks, and they cannot be confined within a single bounding box. 

\textit{FinnWoodlands} comprises three "stuff" categories and five "things" categories. The "stuff" categories are "Lake", "Ground", and "Track", where "Track" refers to a walking trail or path. The "things" categories are "Obstacle", "Spruce", "Birch", "Pine", and "Tree". The class "Obstacle" refers to obstacles on or near a walking trail. "Spruce", "Birch", and "Pine" refer to the tree species which are commonly encountered in Finnish forests. Any other tree that did not fall into these categories was labeled under the general category "Tree". By including "stuff" and "things" categories in our dataset, we aim to provide a more holistic 3D representation of forest scenarios, using panoptic segmentation annotations and sparse depth maps of the scenes.

The vast majority of objects annotated and segmented are tree trunks, accounting for approximately $60.6\%$ of the total amount of objects. Table \ref{classes} shows the overall representation of classes. Within the tree species, "Spruce" is the most common type of tree encountered in our dataset, representing approximately $32.5\%$ of the total amount of objects annotated in our dataset, as shown in Table \ref{tree_classes}.

\begin{figure}[!ht]
\centering
\begin{tabular}{cccc}

{{\rotatebox[origin=l]{90}{\thead{RGB }}}}&
{\includegraphics[width = 1.4in]{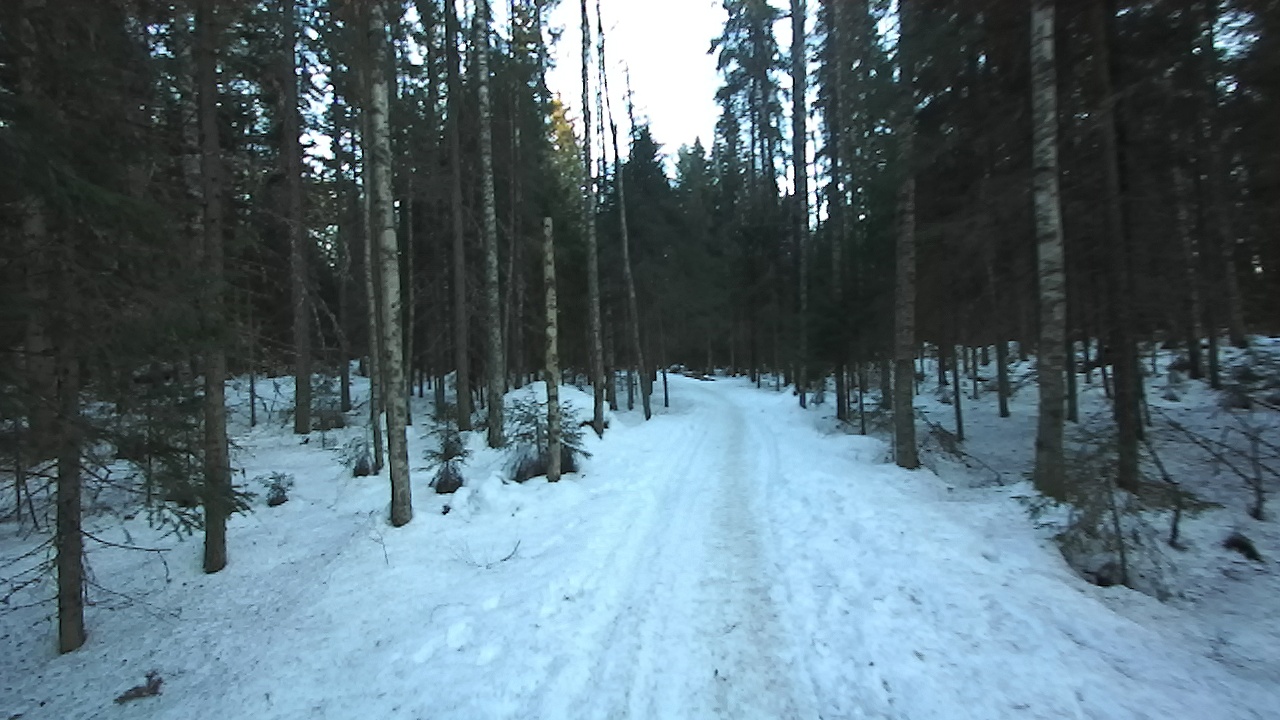}} &
{\includegraphics[width = 1.4in]{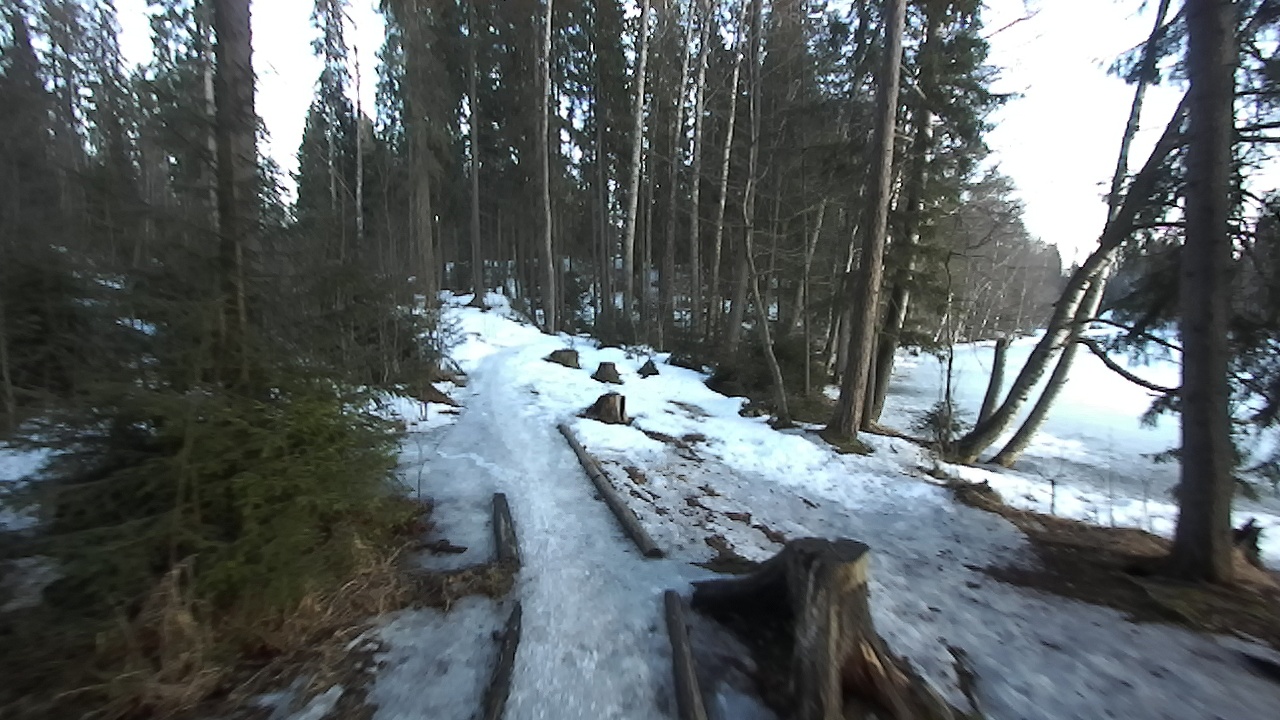}} &
{\includegraphics[width = 1.4in]{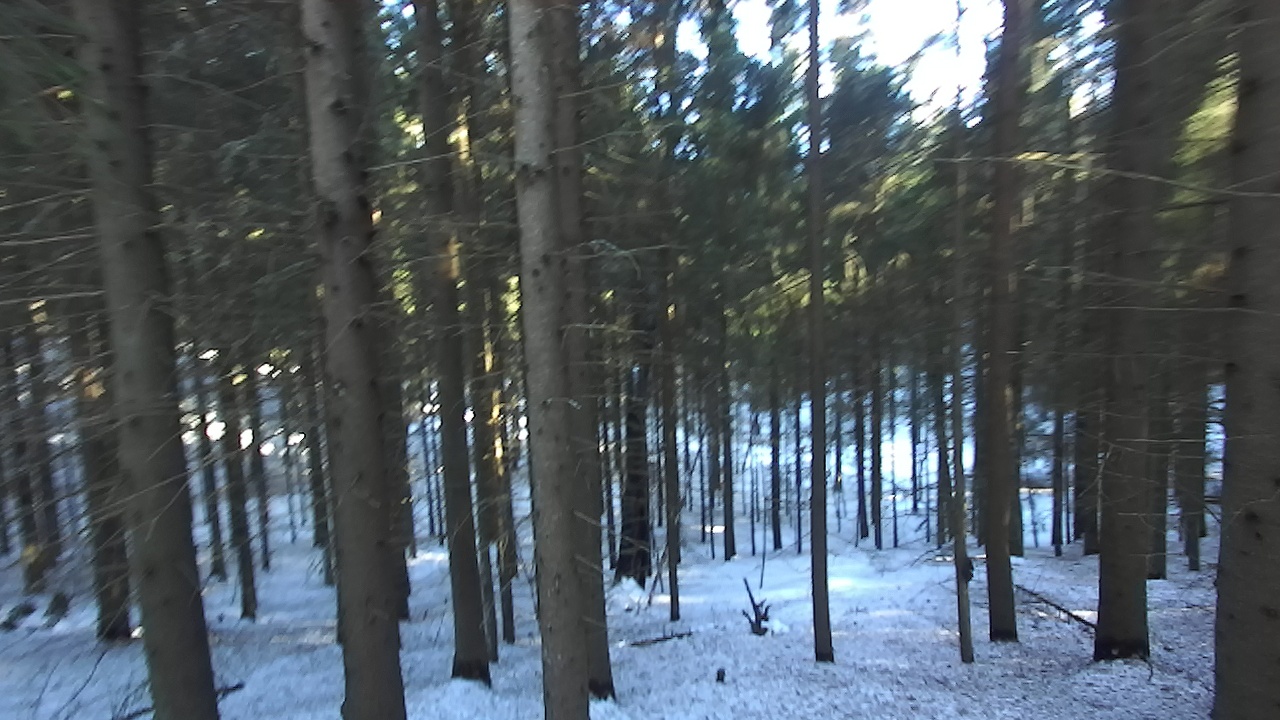}} \\

{{\rotatebox[origin=lB]{90}{\thead{Semantic \\ Segm.}}}} &
{\includegraphics[width = 1.4in]{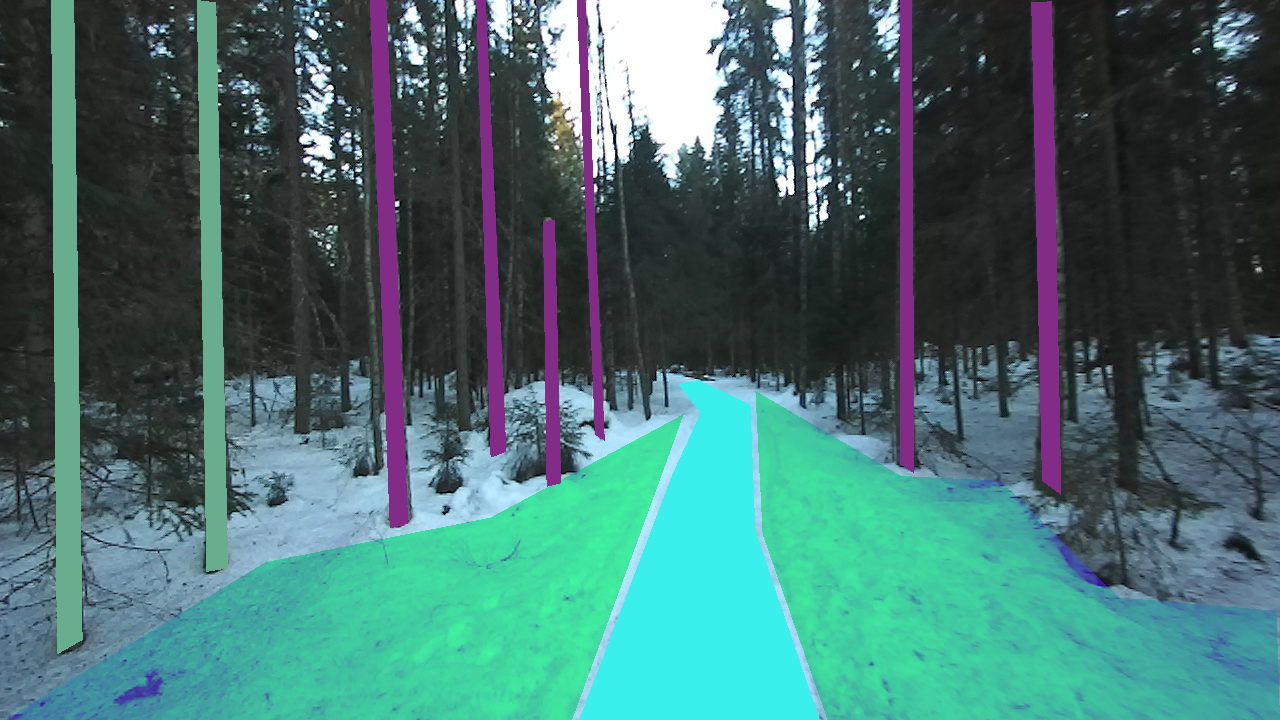}} &
{\includegraphics[width = 1.4in]{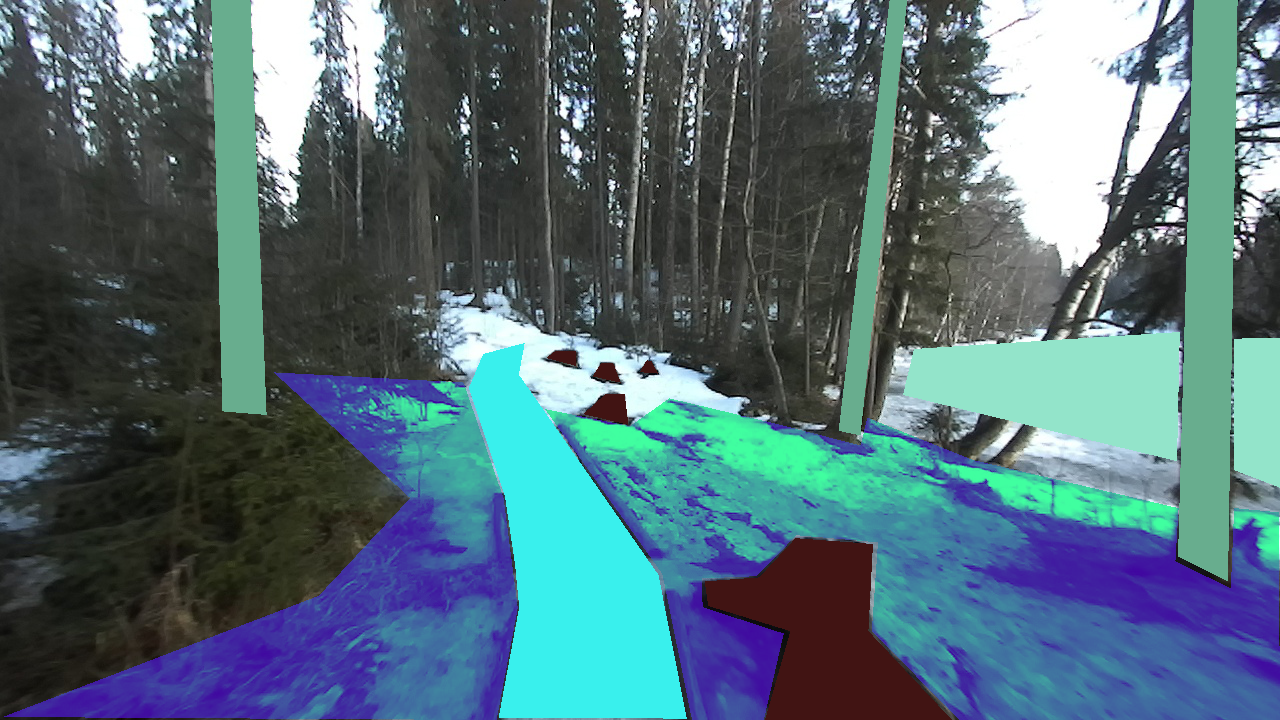}} &
{\includegraphics[width = 1.4in]{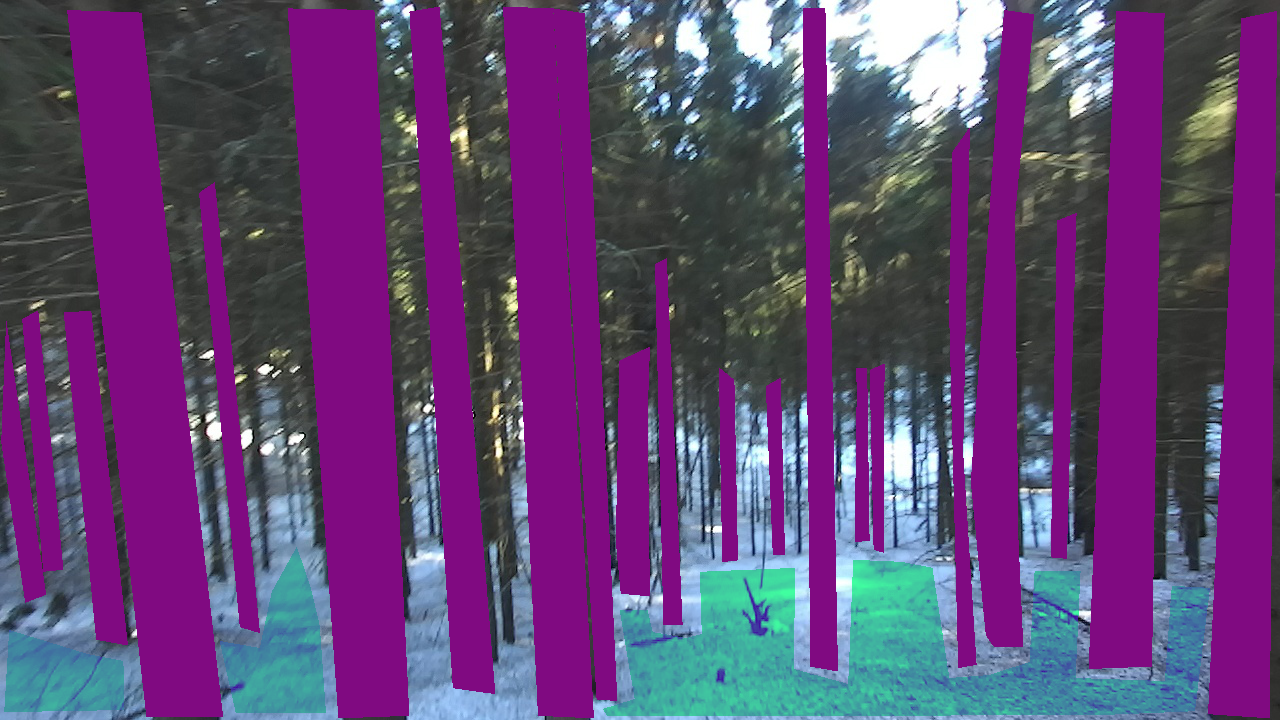}}\\

{{\rotatebox[origin=lB]{90}{\thead{Instance \\ Segm.}}}}&
{\includegraphics[width = 1.4in]{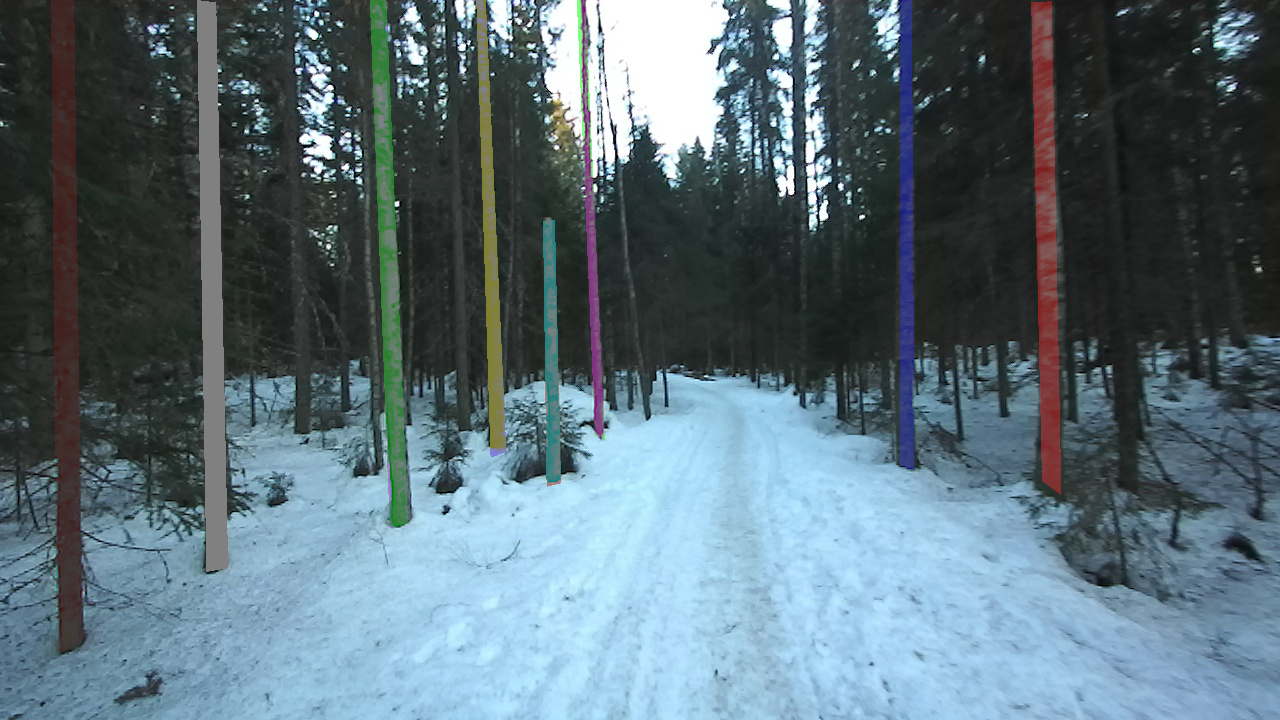}} &
{\includegraphics[width = 1.4in]{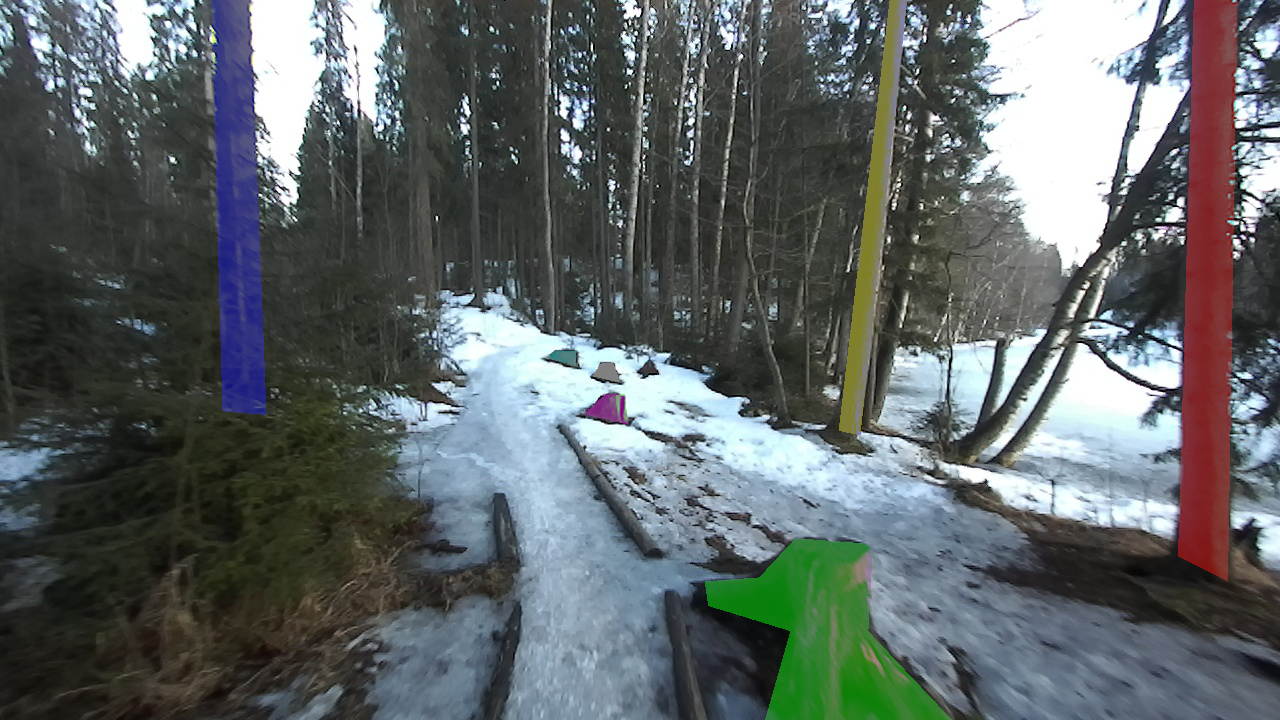}} &
{\includegraphics[width = 1.4in]{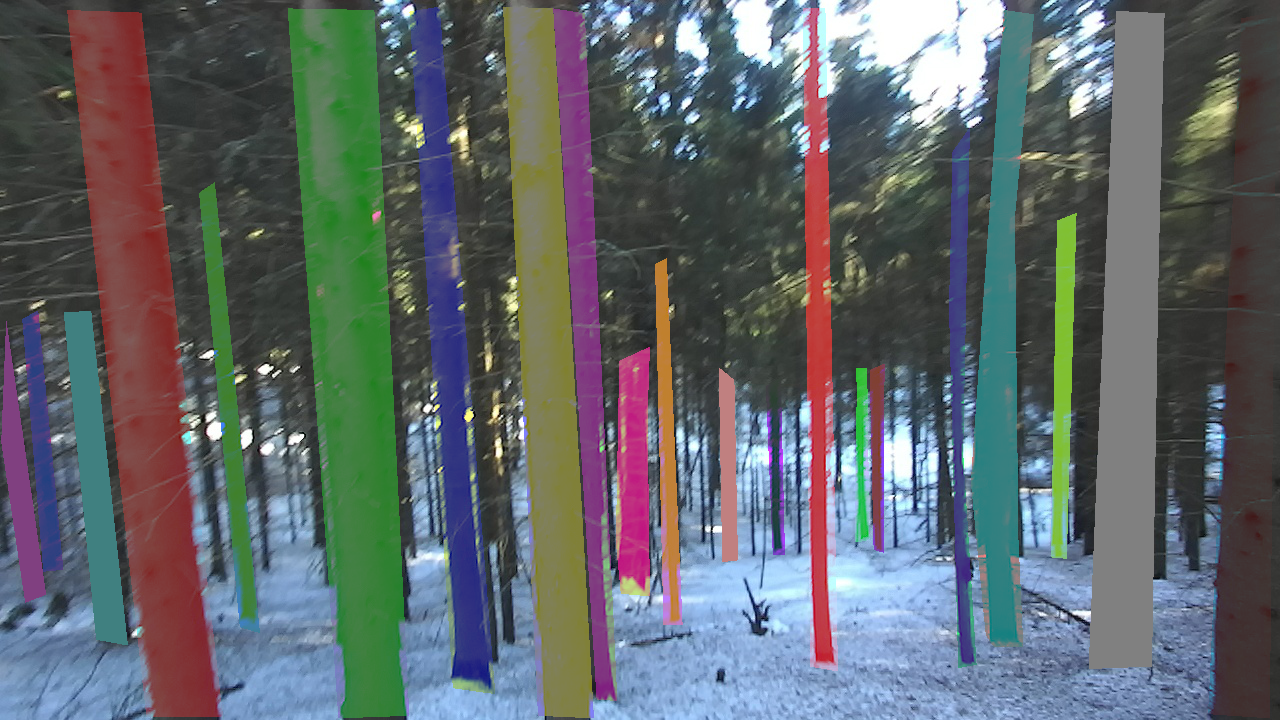}} \\

{{\rotatebox[origin=lB]{90}{\thead{Panoptic \\ Segm.}}}}&
{\includegraphics[width = 1.4in]{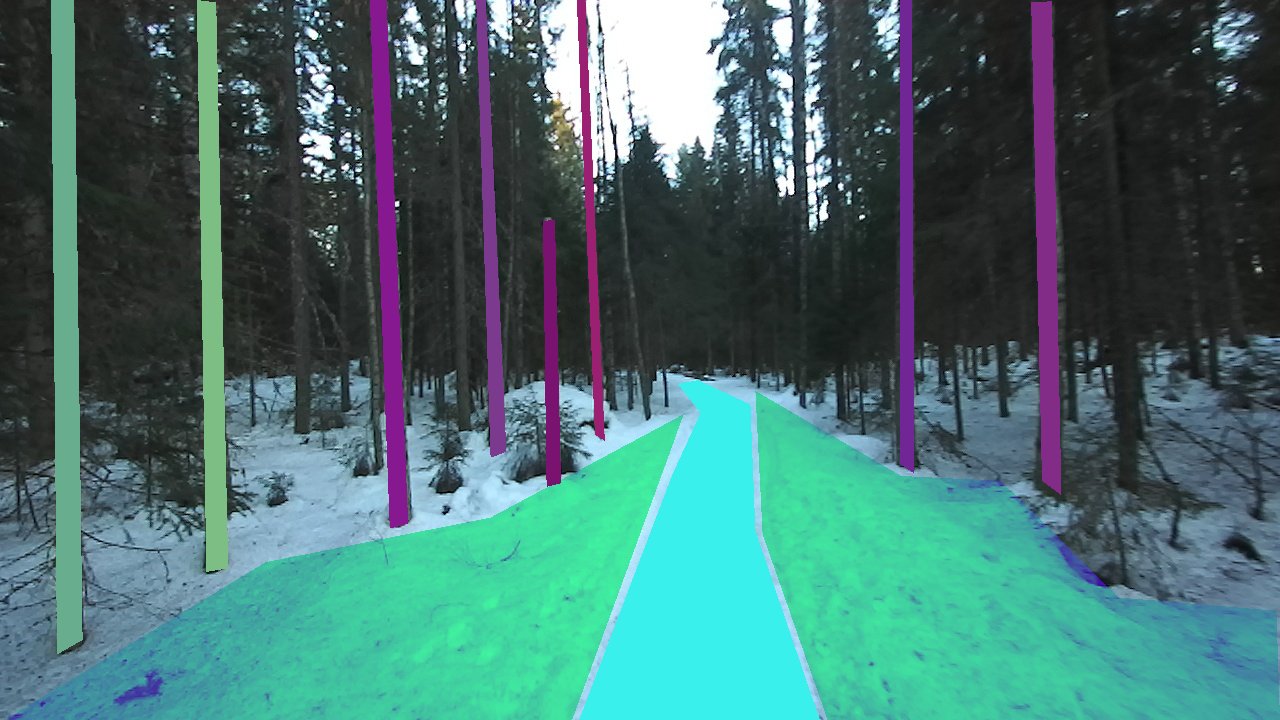}} & 
{\includegraphics[width = 1.4in]{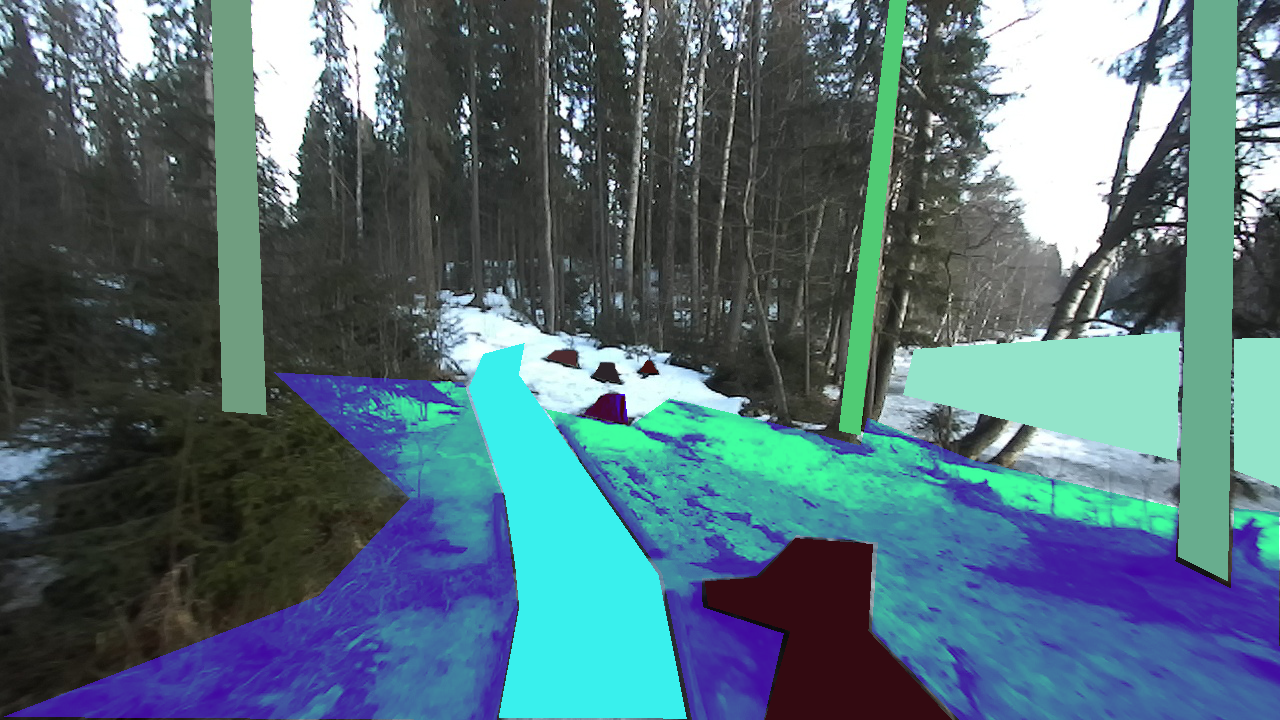}} & 
{\includegraphics[width = 1.4in]{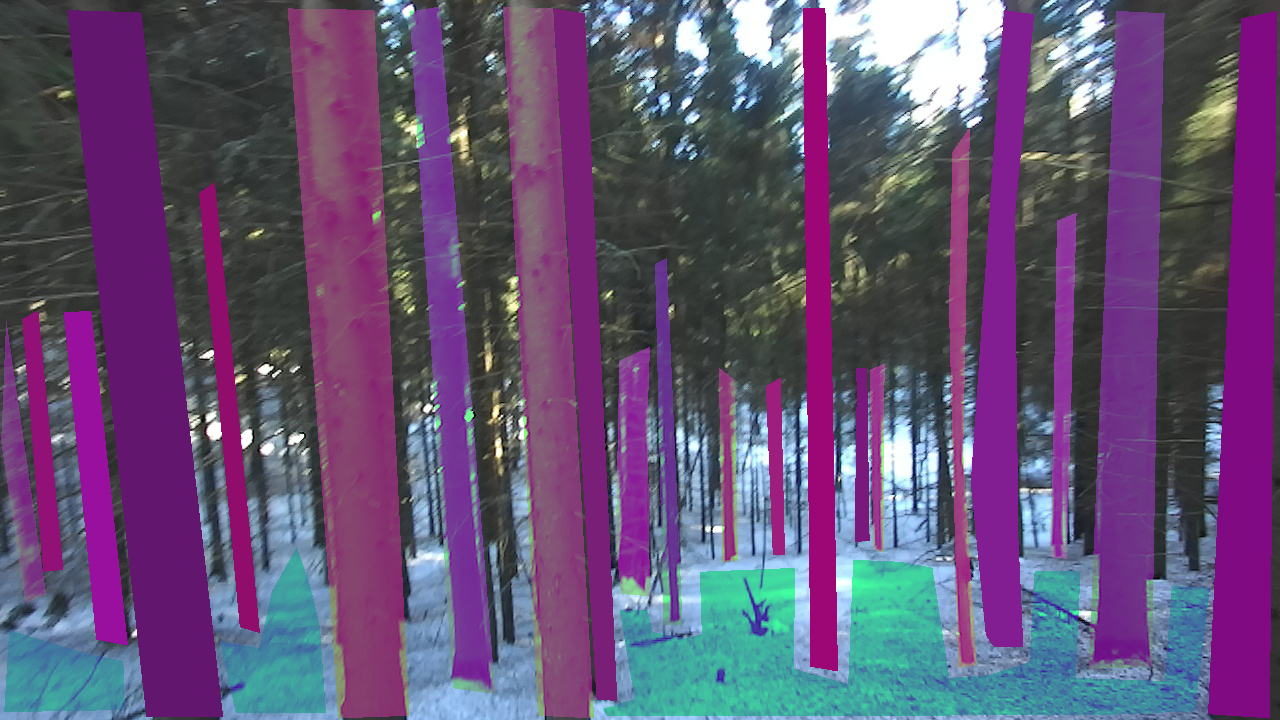}} \\

\end{tabular}
\caption{ FinnWoodlands GT Annotations. Representative samples of our dataset \textit{FinnWoodlands}. The first row displays three different RGB images. Gt annotations for semantic segmentation, instance segmentation and panoptic segmentation are shown from the second to the fourth row respectively.}
\label{finn_ann}
\end{figure}

\begin{table}[h!]
\centering
\caption{FinnWoodlands Object Classes}\label{classes}
 \begin{tabular}{||c c c c||} 
 \hline
 Object Class &  Total Count & Total Area(px) & Representation\\ [0.5ex] 
 \hline\hline
Lake & 378  & 8.728M & 8.9\% \\
Obstacle & 525  & 0.907M  & 12.4\% \\
Ground & 554 & 73.873M & 13.1\%  \\
Track &  207 & 7.778M &  4.8\%  \\
Tree &  2562 & 27.906M & 60.6\%   \\[1ex] 
 \hline
 \end{tabular}
\end{table}

\begin{table}[h!]
\centering
\caption{FinnWoodlands Tree Object Classes}\label{tree_classes}
 \begin{tabular}{||c c c c||} 
 \hline
 Tree Type &  Total Count & Total Area(px) & Representation\\ [0.5ex] 
 \hline\hline
Spruce &  1374 & 16.439M & 32.5\%  \\
Birch &  683 & 6.884M &  16.1\% \\
Pine & 430 & 4.262M &  10.1\%  \\  
Tree (other type) &  75 & 0.319M & 1.7\%  \\ [1ex]
 \hline
 \end{tabular}
\end{table}

The annotations can be inspected visually in Figure \ref{finn_ann}, where representative samples of \textit{FinnWoodlands} dataset have been chosen to depict GT images. Every column corresponds to a different scene; the rows show the RGB frames and corresponding annotations. The first-row show three different scenes; the left-most frame is a typical Finnish forest with a walking trail. The frame in the middle column depicts a forest with a walking trail, obstacles, and a lake on the right side. Finally, the rightmost column shows a forest with no walking trail and with a high density of spruce trees. From the second to the fourth row, the corresponding semantic segmentation GT, instance segmentation GT, and panoptic segmentation GT are shown.

\section{Data Collection}

\begin{figure}[!h]
    \centering
    \subfloat[Backpack]{%
        \includegraphics[width=1.5in, height=2in]{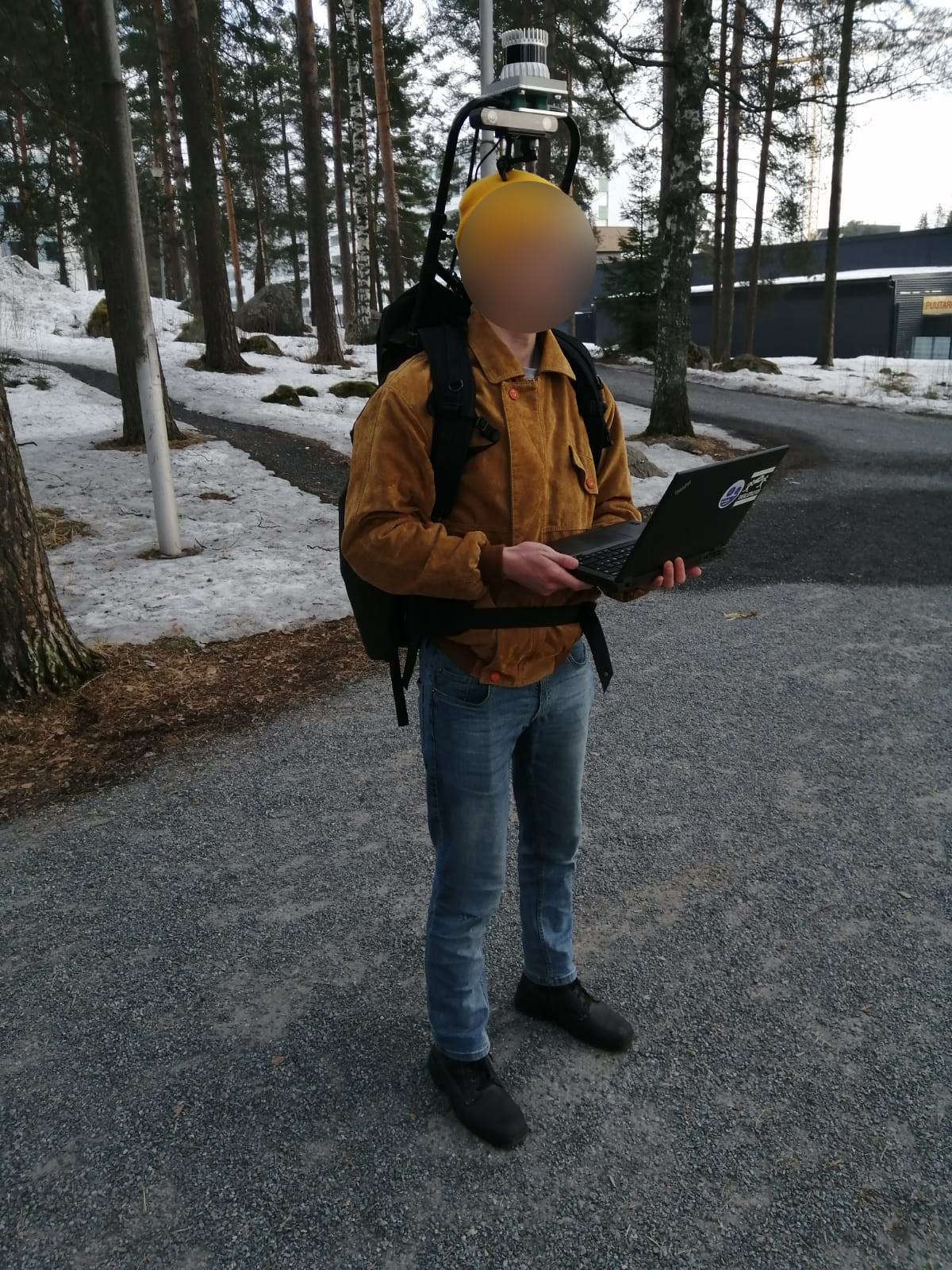}
    }
    \subfloat[Setup]{%
        \includegraphics[width=2.8in, height=2in]{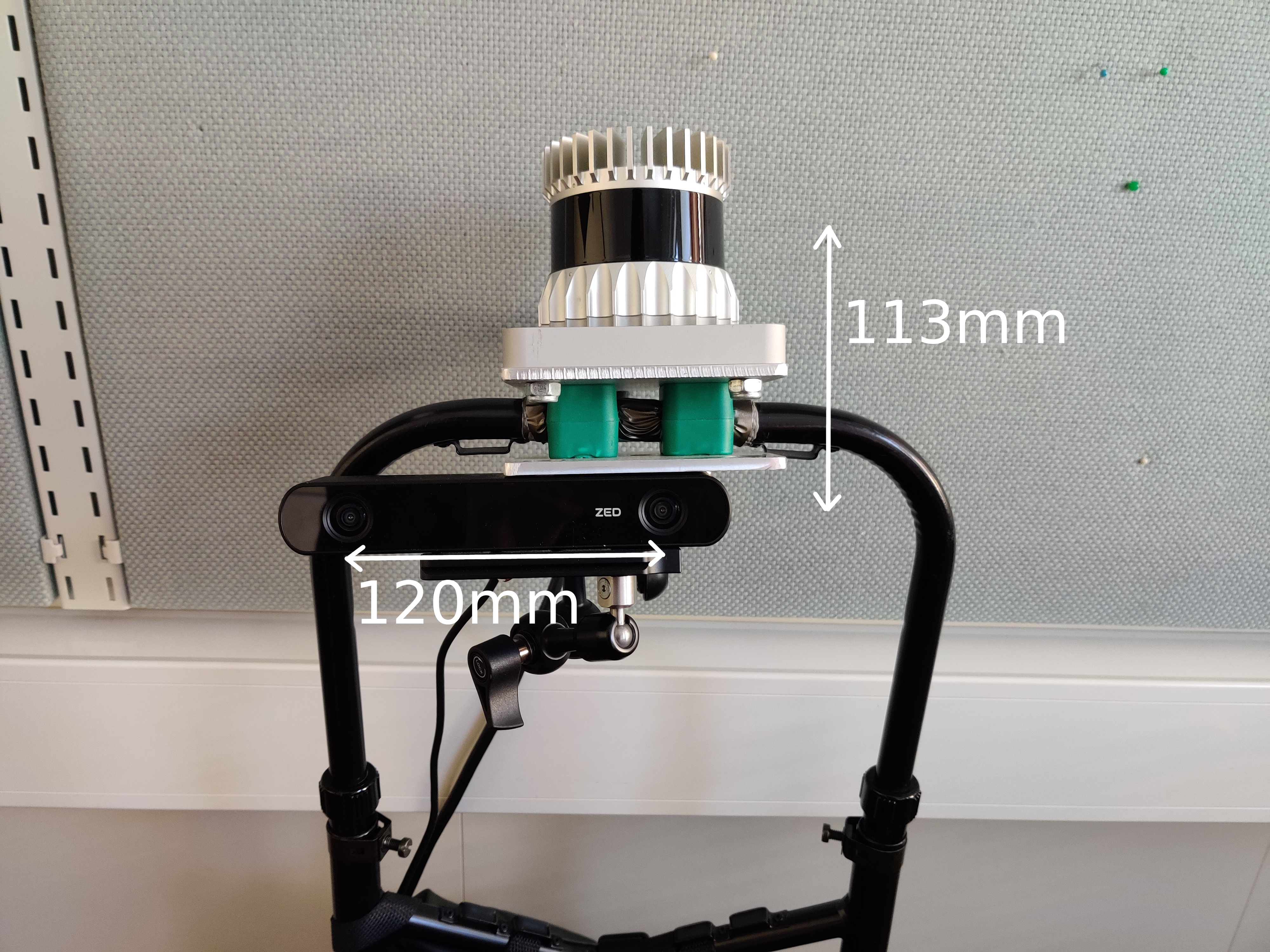}
    }
    \caption{The backpack worn by a user and a close up of the setup. An OS1 LIDAR and a ZED2 stereo camera are mounted on top of a backpack. The ZED2 camera is placed right bellow the LIDAR, at a distance of $113mm$ from the vertical center of the LIDAR. Both the LIDAR and the ZED2 camera are placed above the head of the user using an adjustable metal chassis. The baseline of the ZED2 camera is $120mm$.}
    \label{fig:backpack}
\end{figure}

For the \textit{FinnWoodlands} dataset, a 64-beam Ouster OS1 LIDAR sensor and a ZED2 stereo camera were used to record point clouds and 720p resolution stereo images at 10 frames per second so that for each frame, there are stereo RGB images and one point cloud.
The LIDAR and the camera are securely mounted on the backpack on a metal chassis, they are on top of the wearers head with OS1 above and ZED2 right below, as seen in Fig.\ref{fig:backpack}.
One Lenovo ThinkPad T440P laptop was used to collect and synchronize data from the cameras and sensor. The computer was running Ubuntu Linux (64-bit, version: 18.04), Robot Operating System (ROS, version: Melodic), Nvidia CUDA toolkit (version: 10.2), and Linuxptp software. PTP tools provided by linuxptp were used as a master clock for synchronizing the sensors. The data was then recorded using the ROS tool.

Recording sessions lasted about 50-100 seconds each. They were done by walking at a leisurely pace on a trekking path or in the forest. The locations were chosen from different types of forests near Tampere, Finland. These locations are Kyötikkälä, Hervantajärvi, and Suolijärvi. There is snow on the ground, and the lake is frozen since the recording sessions took place in the early spring. From the three locations, Hervantajärvi and Suolijärvi are surrounded by lakes with more birch trees, whereas Kyötikkälä has spruce trees primarily. 

Forests are a difficult environment to work in compared to urban scenarios. The ground is not even, and there are many obstacles, e.g. rocks, trees, pits, and hills. Some places proved to be challenging to collect data by walking due to the snow on the ground and other obstacles. However, our data-collection setup, albeit simple, is flexible enough to collect reliable data under such conditions.

\section{Experiments}

In forestry applications, as in other applications, it is essential to understand the semantic meaning of a given image. Very often interacting with the objects present in the scene is also necessary. For that matter, semantic segmentation is not enough since it only provides pixel-wise classification, whereas instance segmentation allows for object detection and object segmentation. Moreover, it is possible to combine instance and semantic segmentation using panoptic segmentation, which provides a more holistic representation of the scenes. On the other hand, depth completion provides depth values for every pixel on the input frames, given incomplete sparse depth maps as input, which is relevant, especially for applications where it is needed to interact with objects in a 3D space.

Therefore, we conducted baseline experiments on three different models, namely, Mask R-CNN \cite{maskrcnn} for instance segmentation, EfficientPS \cite{effps} for panoptic segmentation and FuseNet \cite{2d3d} for depth completion based on sparse depth maps and RGB images. Our training set consists of $150$ frames collected from two different locations, and the evaluation set contains $50$ frames collected from a different location not used in our training set.

In order to train the depth completion model \textit{FuseNet} \cite{2d3d}, we sub-sampled the sparse depth maps and used the resulting sparse depth maps as training data while keeping the original sparse depth maps as GT. The sub-sampled sparse depth maps contain $3500$ depth values which account for approximately $25\%$ of points in the original depth maps.

\paragraph{\textbf{Mask R-CNN}}\cite{maskrcnn} reached the state-of-the-art of end-to-end instance segmentation in 2017 and has been broadly used as a reference model ever since. It takes RGB images as input and returns three different types of outputs. More specifically, it returns bounding boxes for object localization, class labels for every object detected, and segmentation masks. It consists of a backbone for feature extraction, a region proposal network (RPN), and three output heads in parallel that predict the corresponding class labels, bounding boxes, and segmentation masks for the given input image. We used Mask R-CNN with EfficientNet-B5 as the backbone \cite{effnet}, pre-trained on the ImageNet dataset \cite{imnet}. We replaced the batch normalization layers \cite{bn} with synchronized Inplace Activated Batch Normalization \cite{ibn} for GPU optimization.

\paragraph{\textbf{EfficientPS}}\cite{effps} is a model for end-to-end panoptic segmentation. Similar to Mask R-CNN, it uses a backbone to extract multiple feature maps, fed to two output branches, and finally, it contains one panoptic fusion module. The first branch performs instance segmentation based on Mask R-CNN, using an RPN and three output heads for predicting bounding boxes, class labels, and segmentation masks. The second branch performs semantic segmentation, and finally, the panoptic fusion module fuses the semantic logits and the instance segmentation mask logits to produce a panoptic segmentation map. The name of the model makes reference to its backbone, which is based on the scalable \textit{EfficientNet} \cite{effnet} architecture, wrapped in a two-way feature pyramid network (FPN) that allows for feature extraction at multiple scales.

\paragraph{\textbf{FuseNet}}\cite{2d3d} uses RGB images, and sparse depth maps to perform end-to-end depth completion. It returns fully dense depth maps as output. FuseNet learns 2D and 3D features jointly using a building block that processes 2D tensors using 2D convolutional layers and 3D points using continuous convolution. The resulting features of the building block are then fused in 2D space.

\subsection{Evaluation}

We used the standard COCO evaluation metrics \cite{coco}. More specifically, we computed the mean Average Precision (mAP) for evaluating instance segmentation, Mean Intersection over Union (mIoU) for semantic segmentation, Panoptic Quality (PQ), Segmentation Quality (SQ), and Recognition Quality (RQ) for panoptic segmentation. We computed the Root Mean Square Error (RMSE) to evaluate the depth completion task. 

\begin{figure}

\centering
\begin{tabular}{cc}
\thead{Instance \\ Segm.} & \thead{Instance \\ Segm. GT} \\
{\includegraphics[width = 2.2in, height=0.9in]{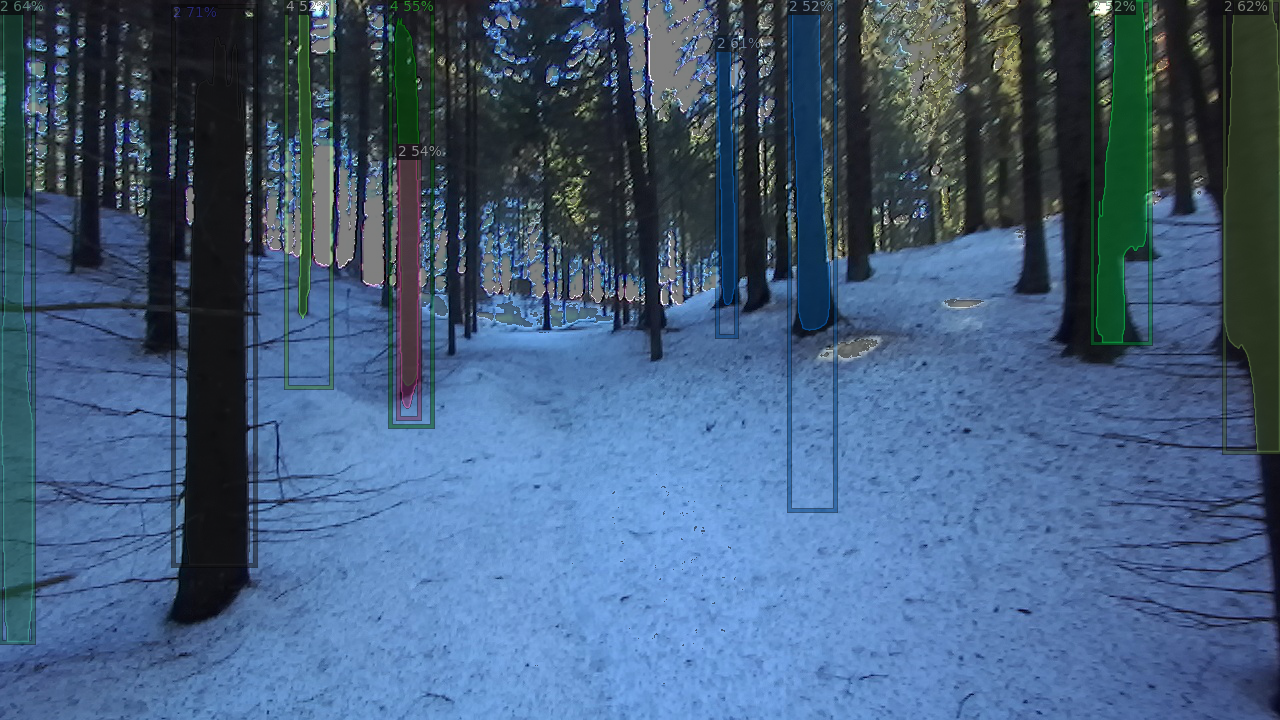}} &
{\includegraphics[width = 2.2in, height=0.9in]{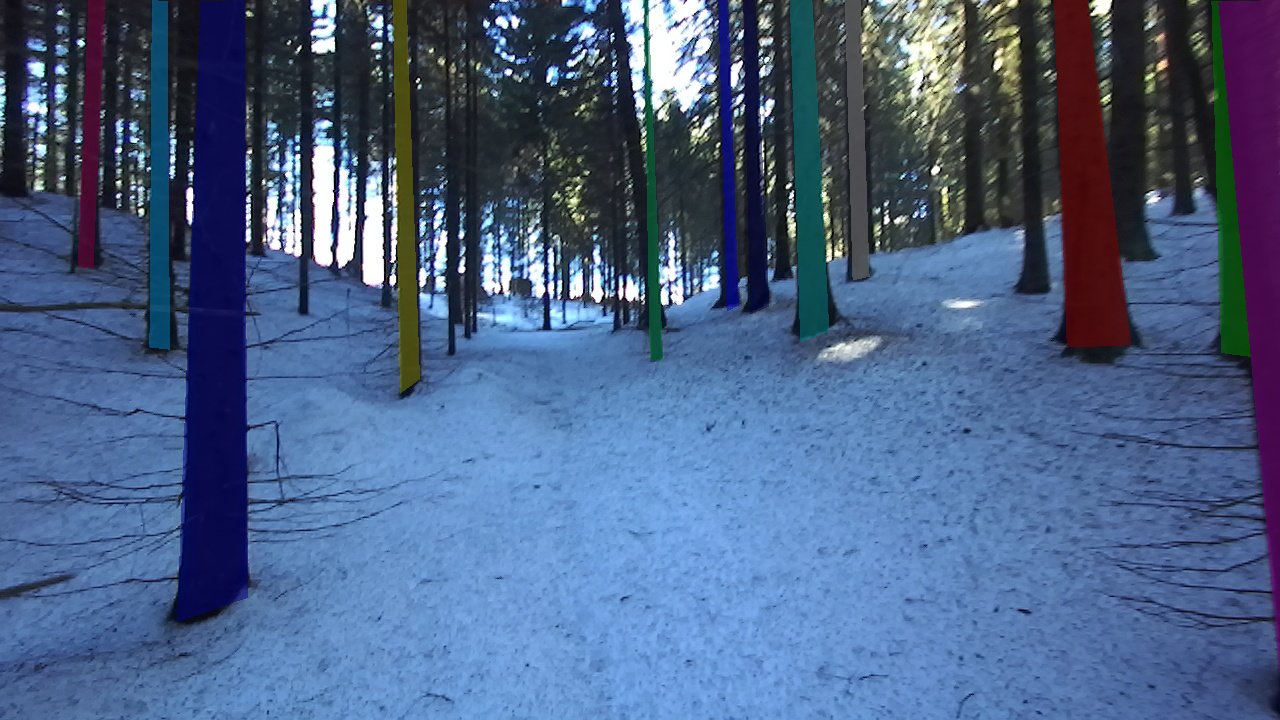}} \\
{\includegraphics[width = 2.2in, height=0.9in]{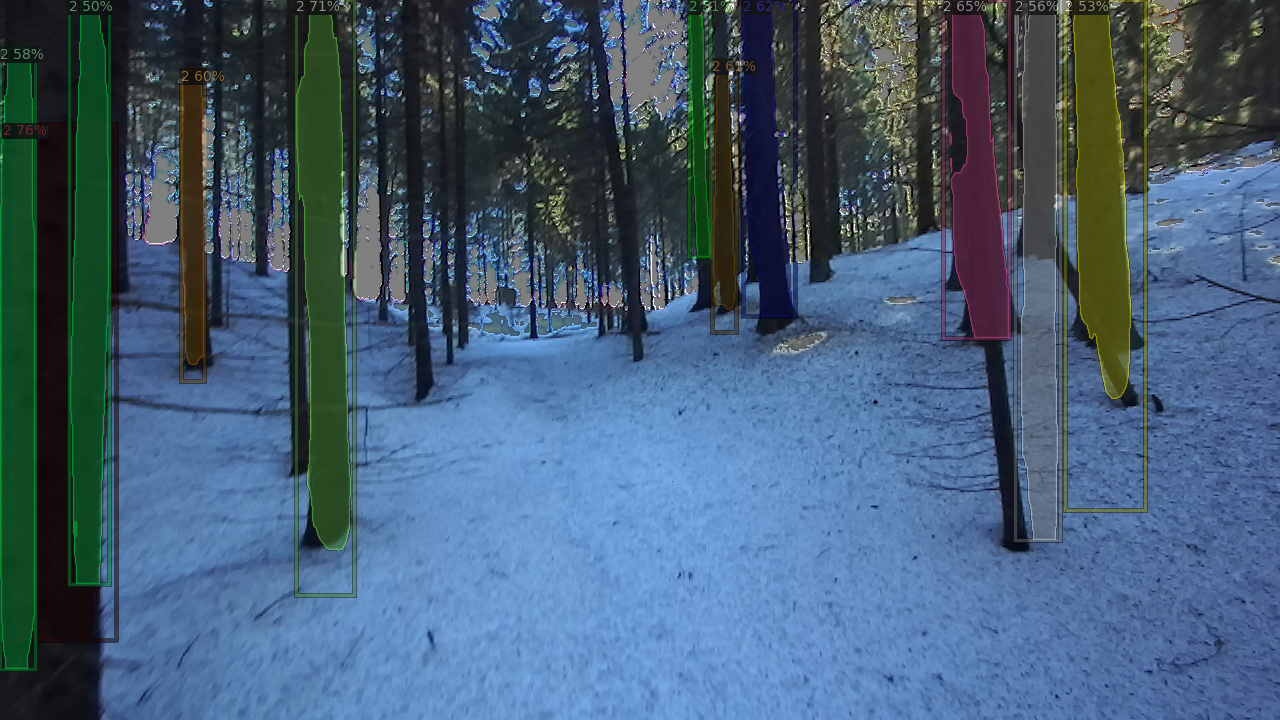}} &
{\includegraphics[width = 2.2in, height=0.9in]{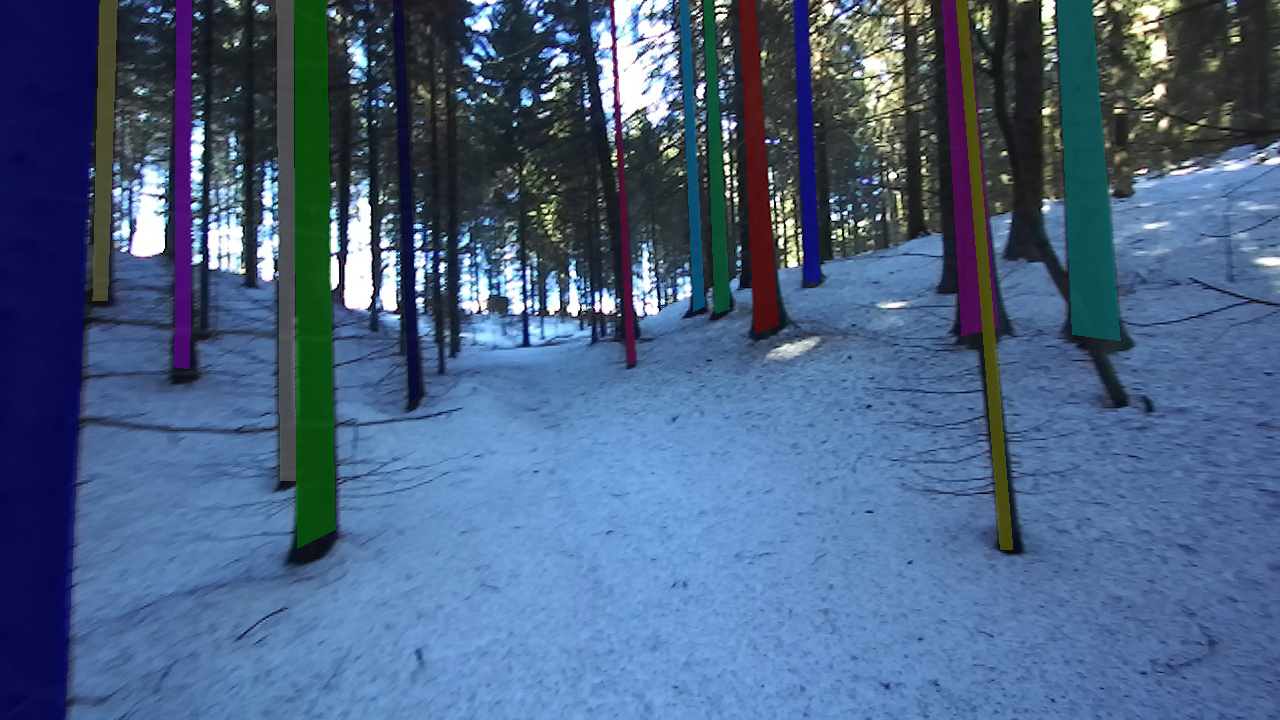}} \\
{\includegraphics[width = 2.2in, height=0.9in]{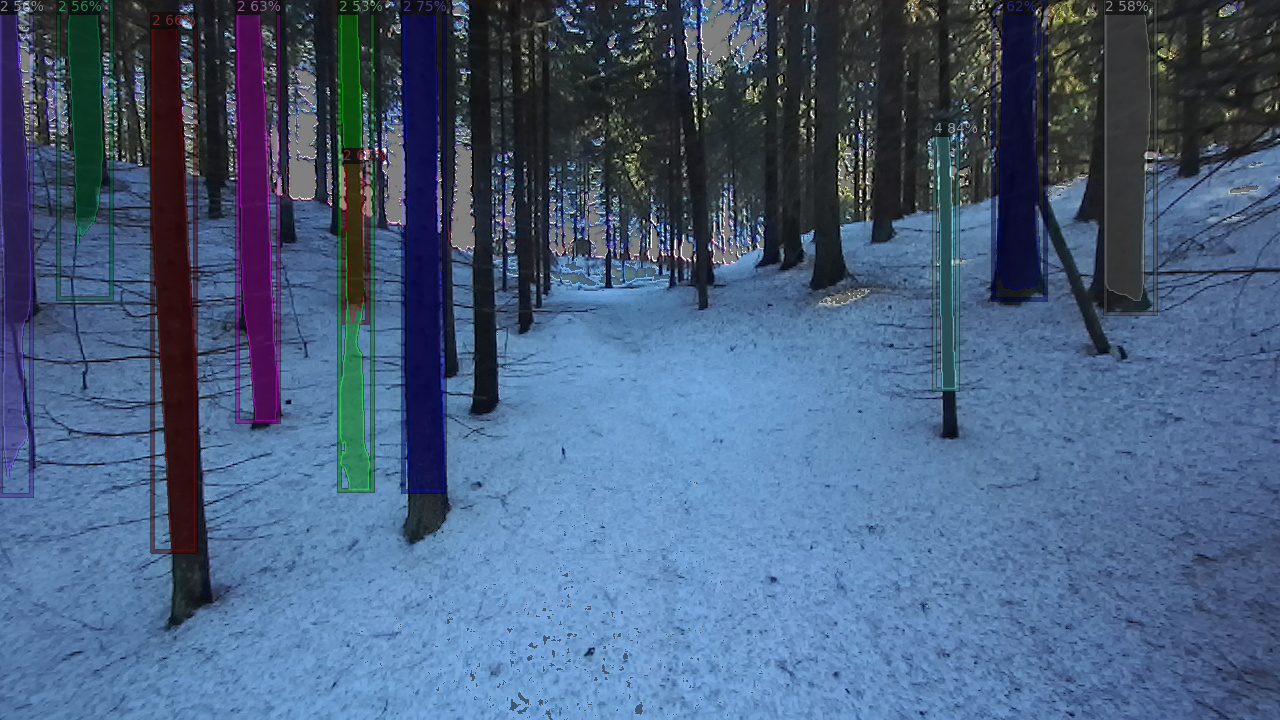}} &
{\includegraphics[width = 2.2in, height=0.9in]{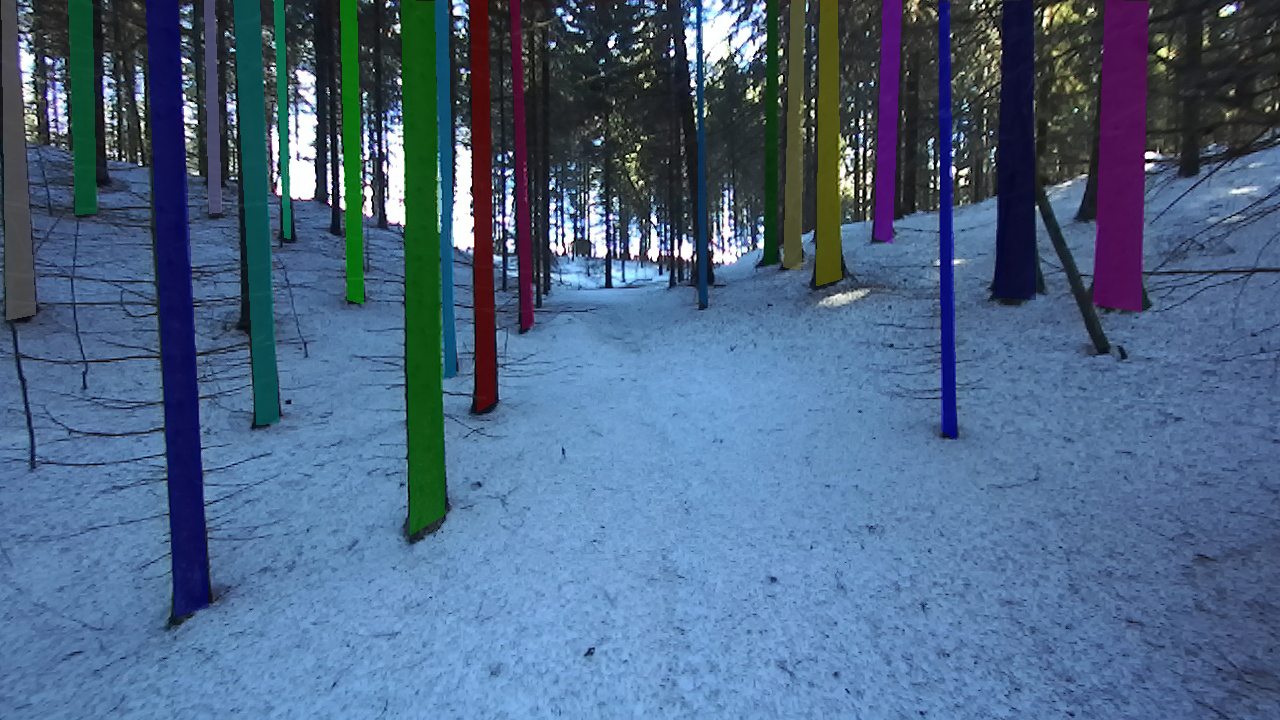}} \\
\end{tabular}
\caption{Mask R-CNN \cite{maskrcnn} Qualitative Results. The first column shows three different input frames. The first and second column, depict the instance segmentation output from Mask R-CNN \cite{maskrcnn} and the GT segmentation respectively.}
\label{mask_res_vis}
\end{figure}

\subsection{Results}

\begin{figure}
\centering
\begin{tabular}{cccc}

{{\rotatebox[origin=l]{90}{\thead{RGB }}}}&
{\includegraphics[width = 1.4in]{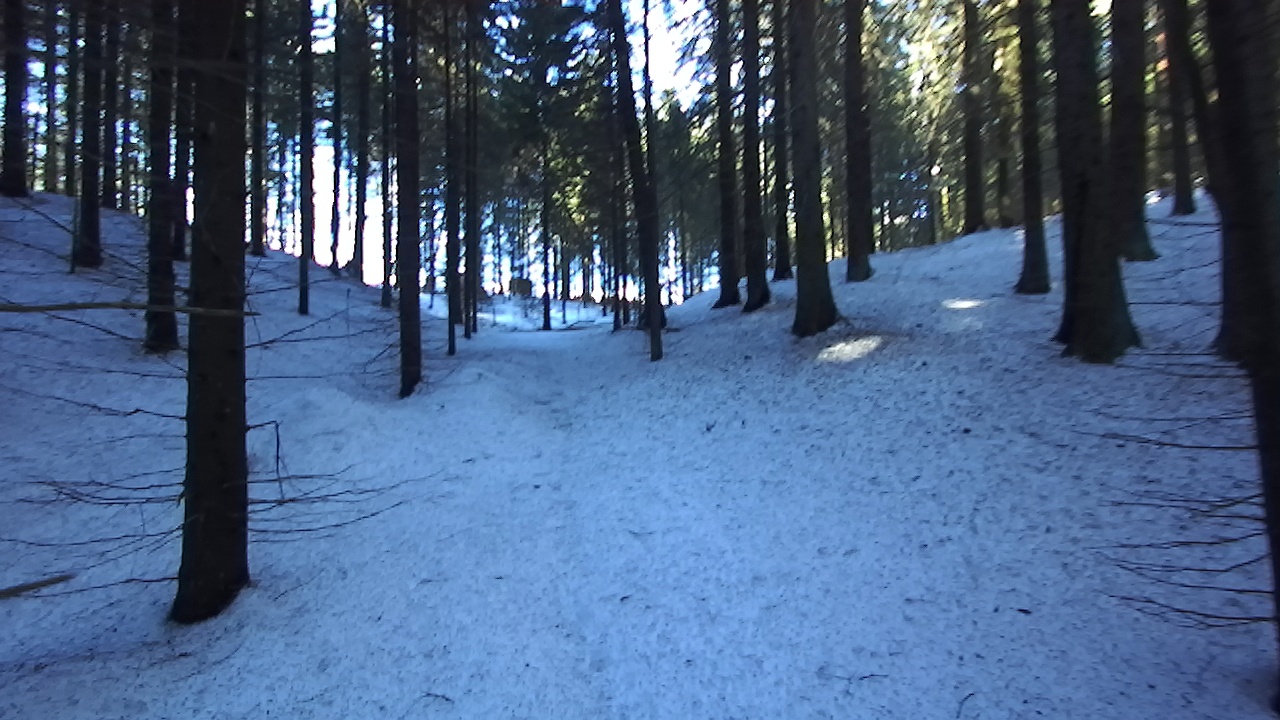}} &
{\includegraphics[width = 1.4in]{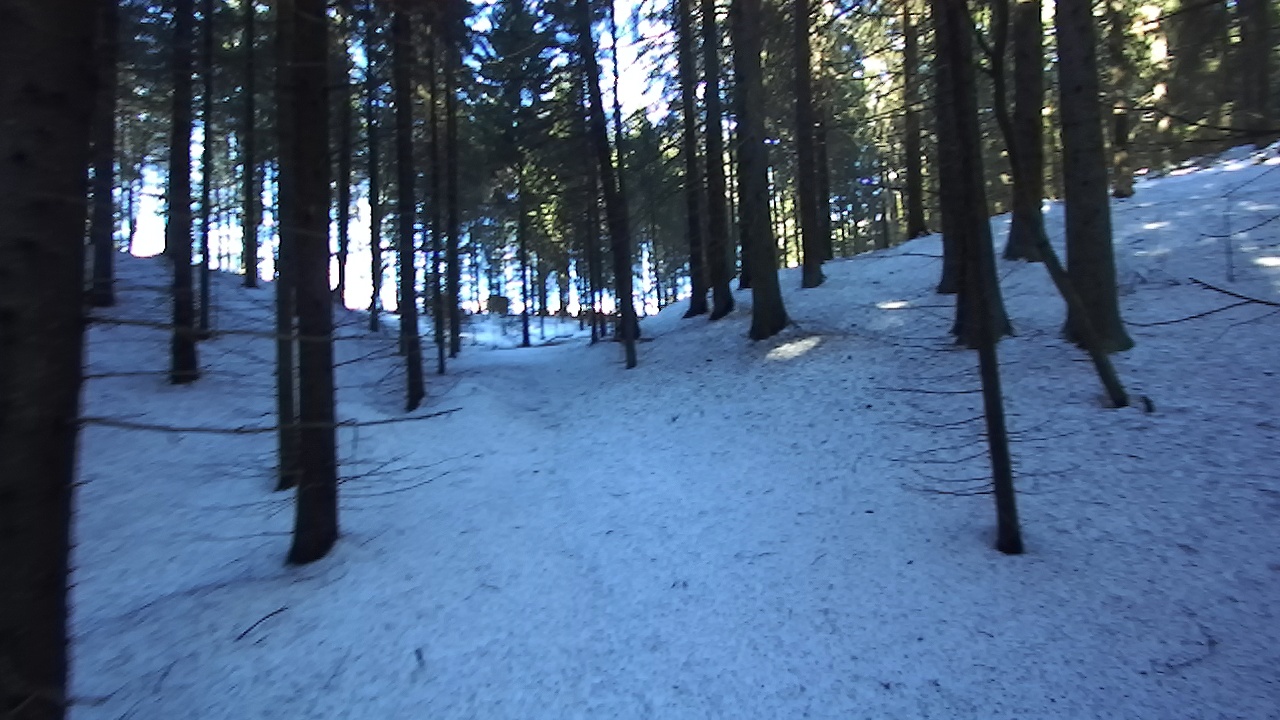}} &
{\includegraphics[width = 1.4in]{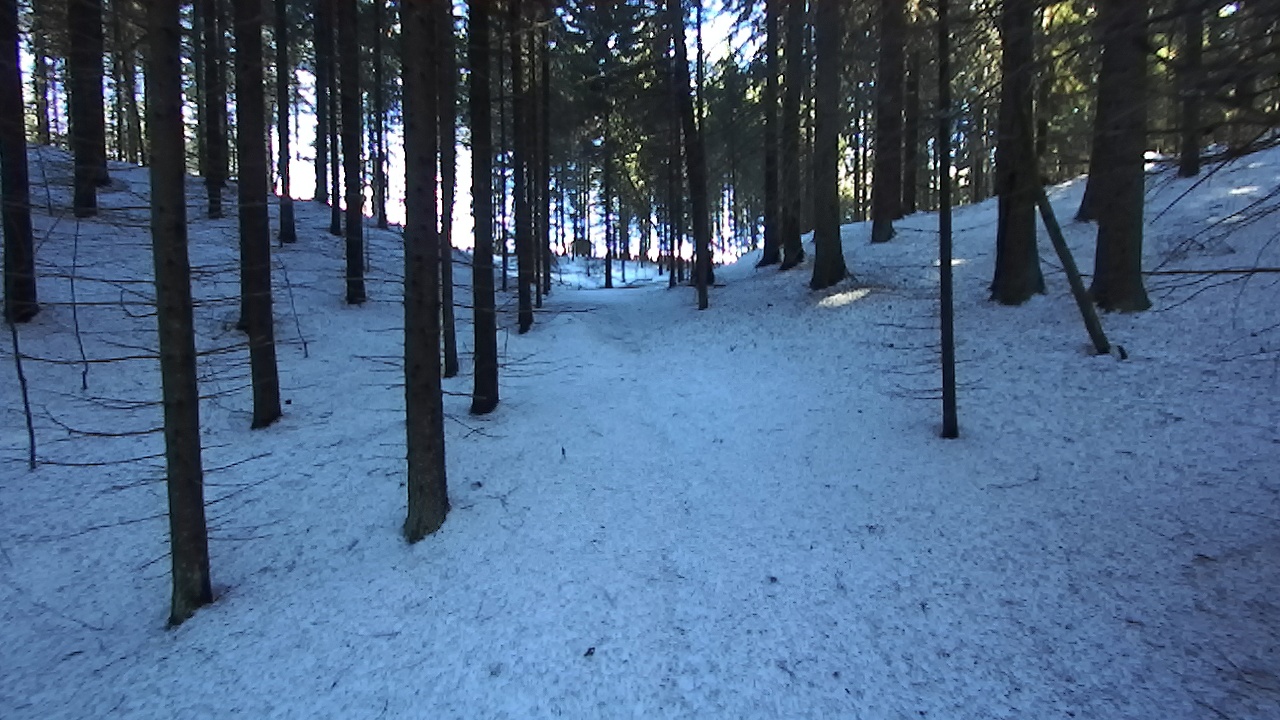}} \\

{{\rotatebox[origin=lB]{90}{\thead{Semantic \\ Segm.}}}} &
{\includegraphics[width = 1.4in]{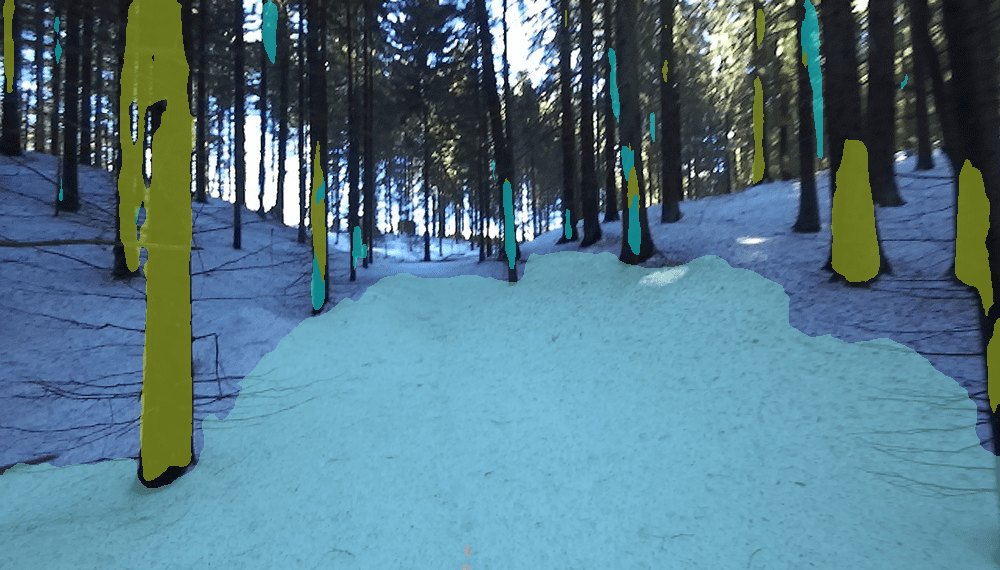}} &
{\includegraphics[width = 1.4in]{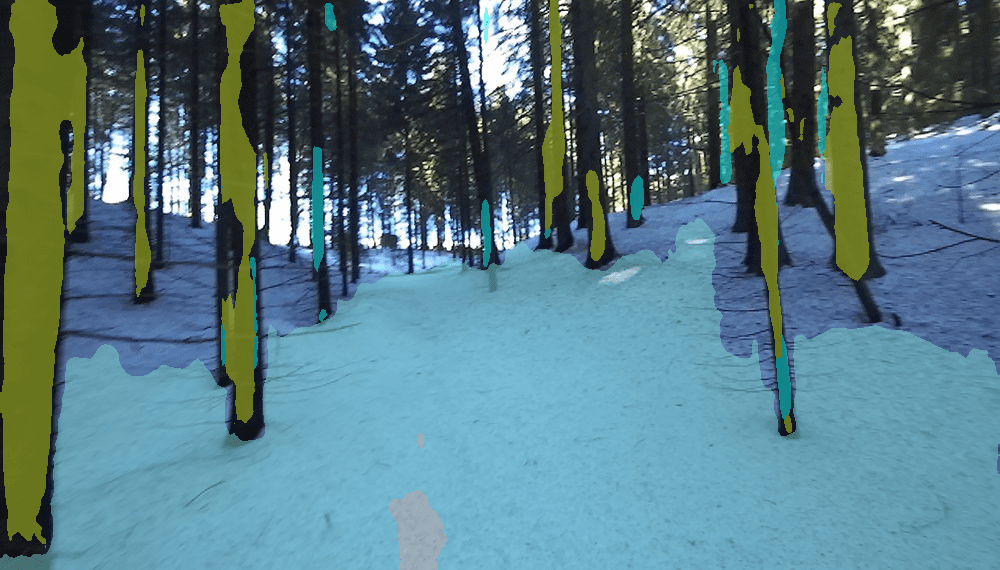}} &
{\includegraphics[width = 1.4in]{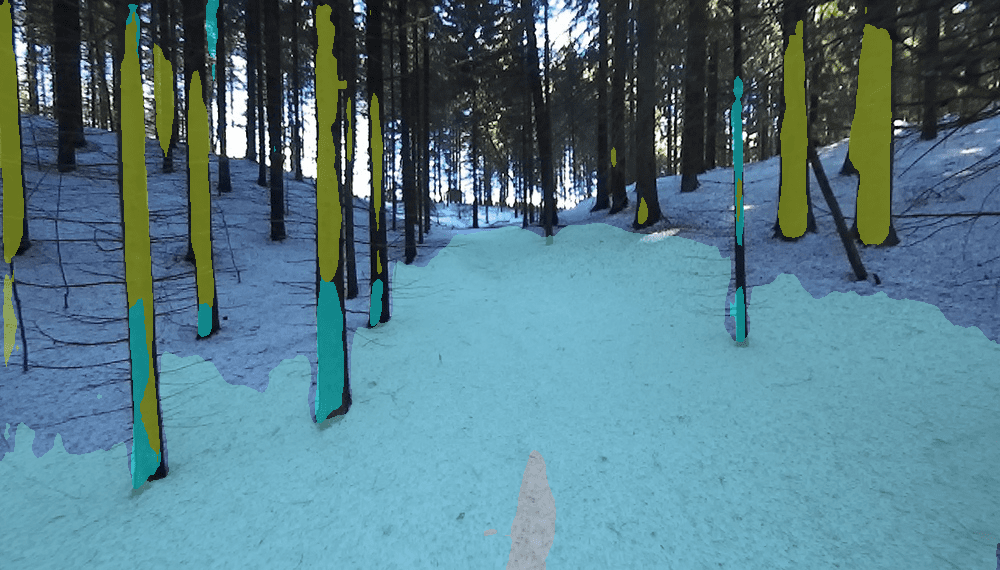}}\\

{{\rotatebox[origin=lB]{90}{\thead{Instance \\ Segm.}}}}&
{\includegraphics[width = 1.4in]{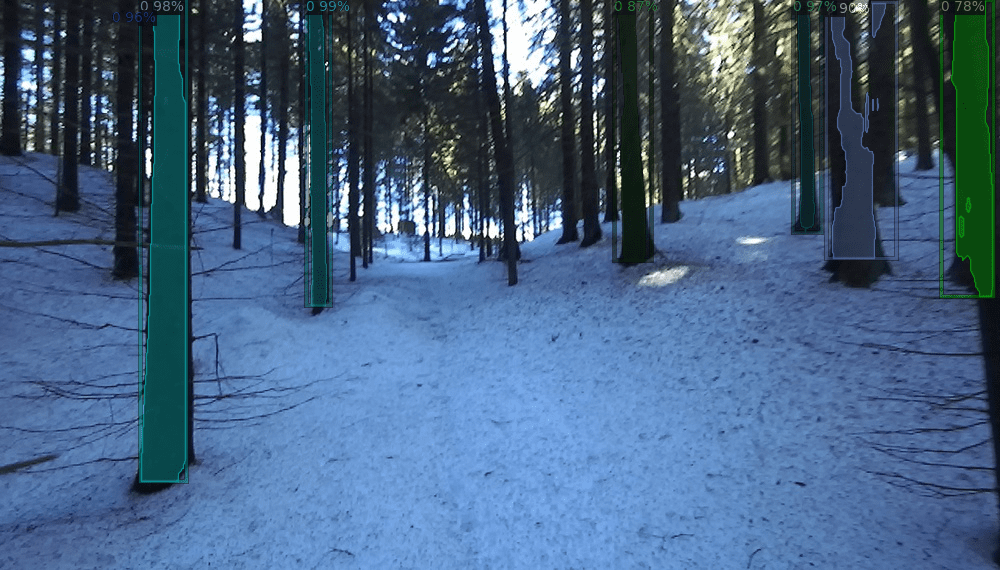}} &
{\includegraphics[width = 1.4in]{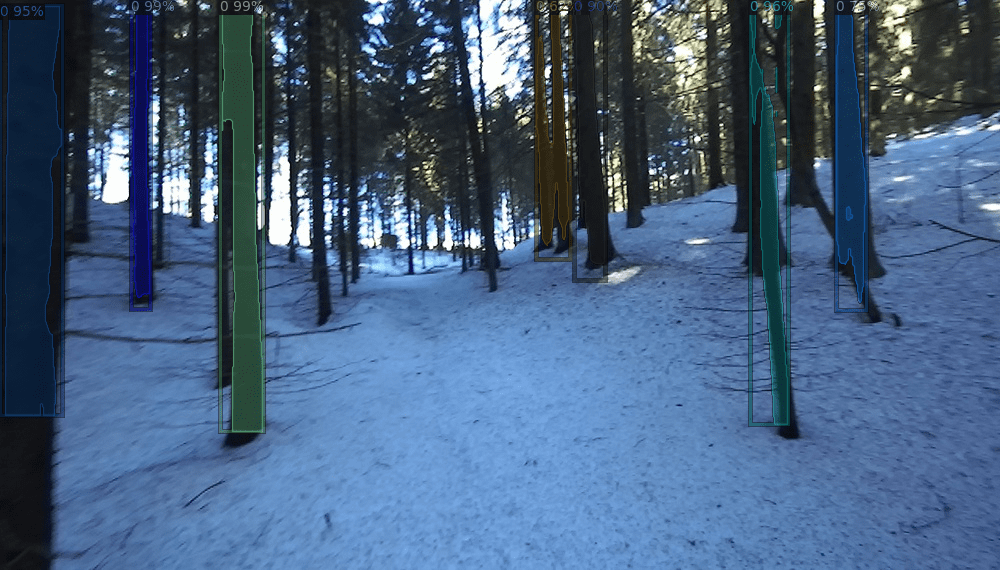}} &
{\includegraphics[width = 1.4in]{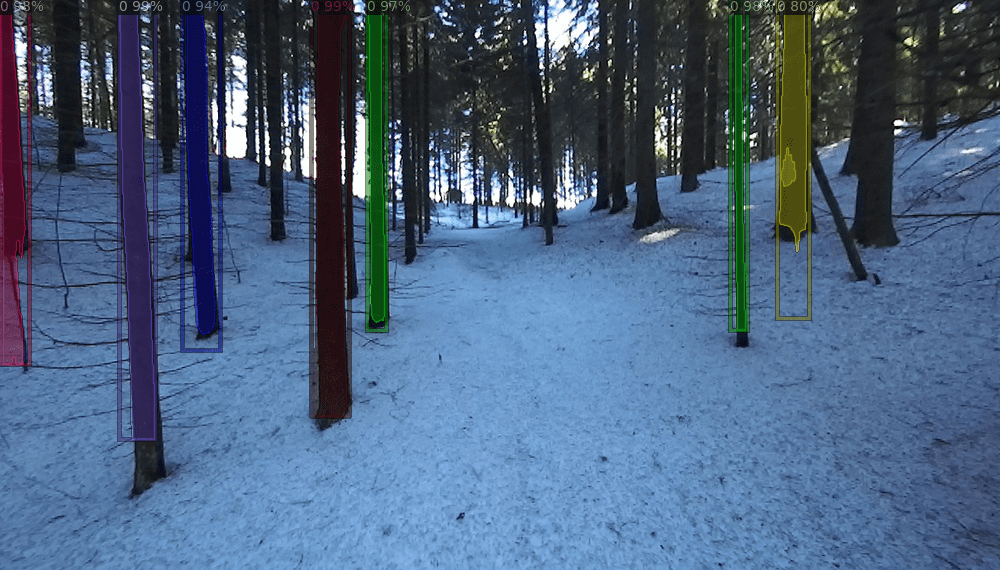}} \\

{{\rotatebox[origin=lB]{90}{\thead{Panoptic \\ Segm.}}}}&
{\includegraphics[width = 1.4in]{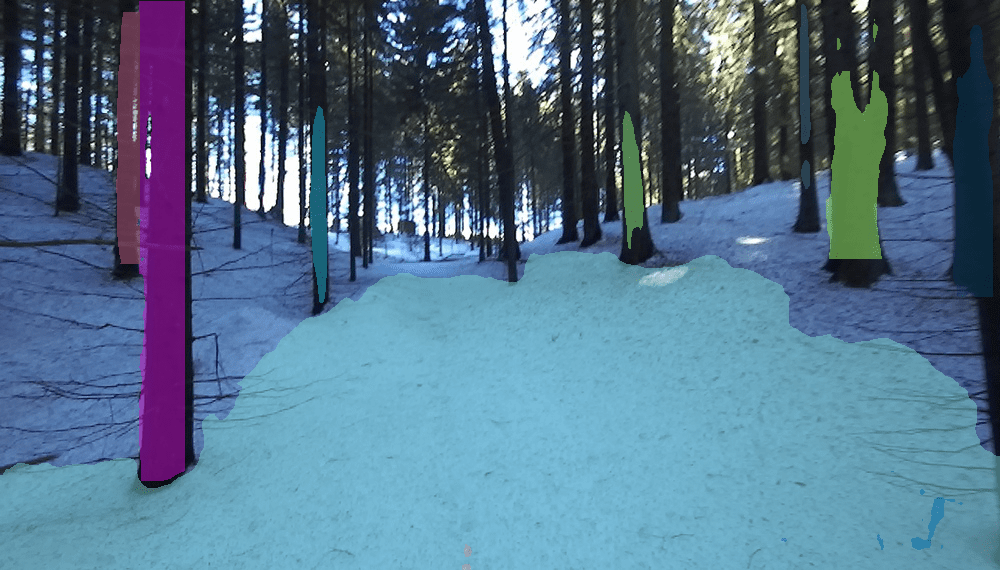}} & 
{\includegraphics[width = 1.4in]{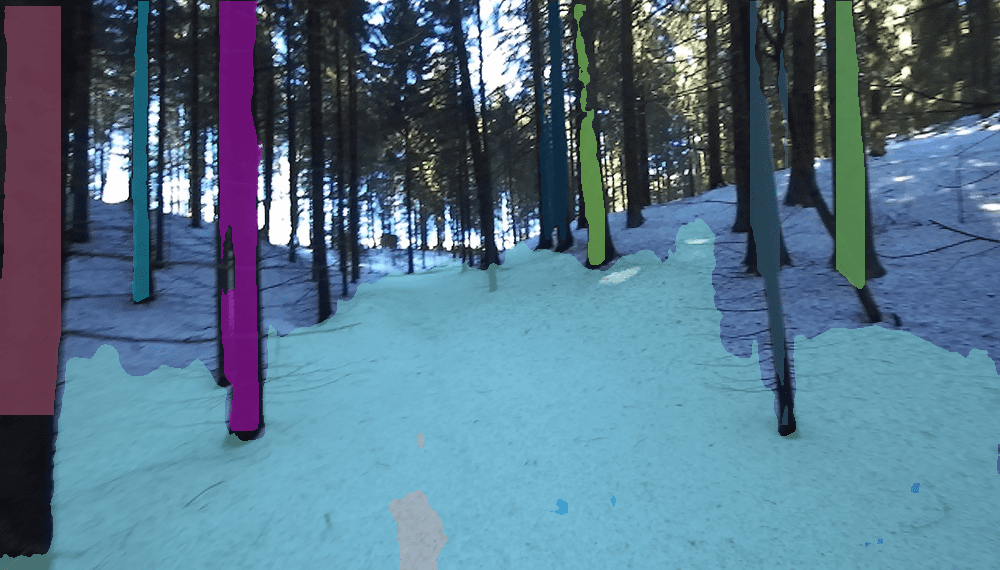}} & 
{\includegraphics[width = 1.4in]{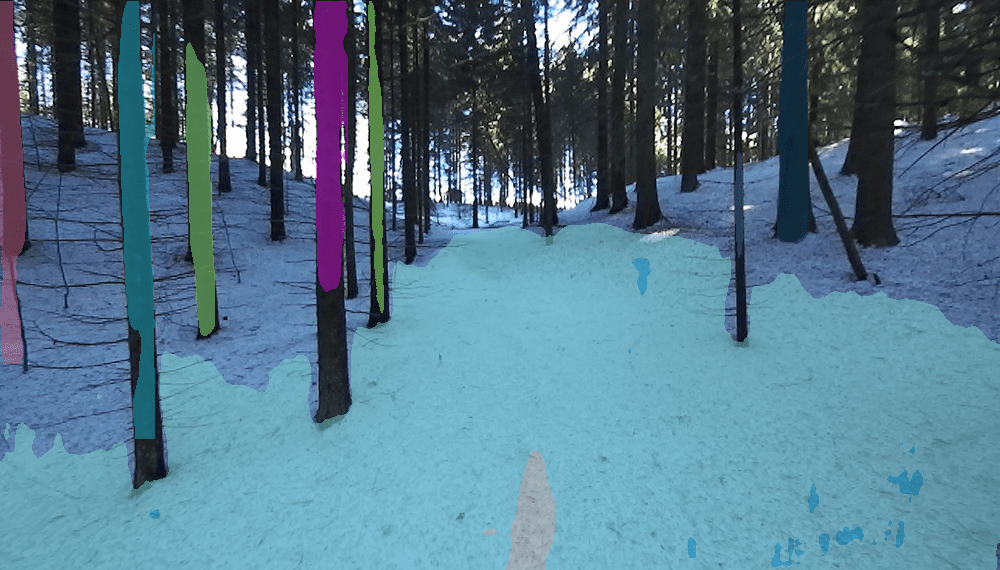}} \\

{{\rotatebox[origin=lB]{90}{\thead{Panoptic \\ GT.}}}}&
{\includegraphics[width = 1.4in]{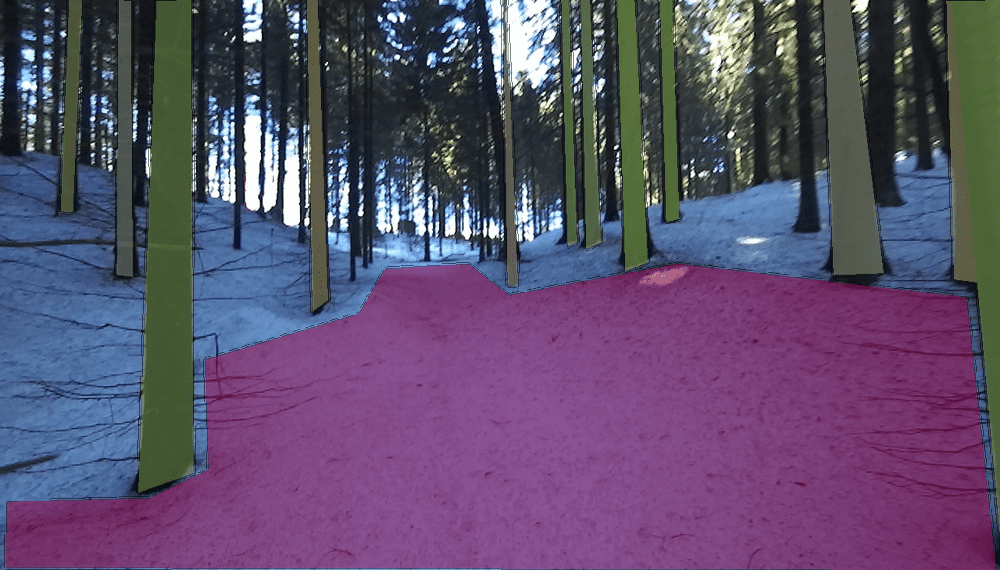}} & 
{\includegraphics[width = 1.4in]{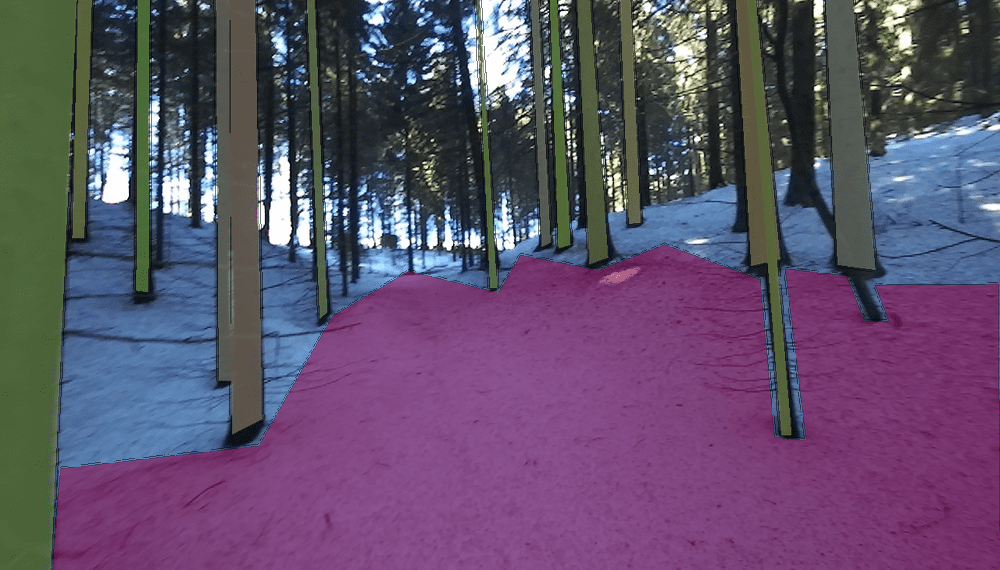}} & 
{\includegraphics[width = 1.4in]{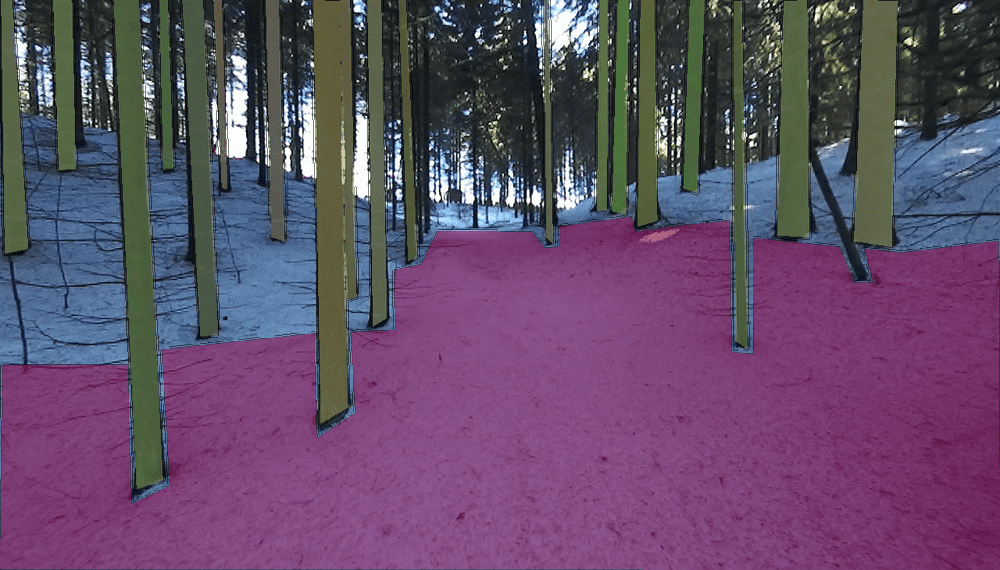}} \\
\end{tabular}
\caption{ EfficientPS \cite{effps} Qualitative Results. On the top row, three different RGB input images are shown. From the second to the fourth row, we present the segmentation results from EfficientPS for every task, more specifically, semantic segmentation, instance segmentation and panoptic segmentation respectively. The last row shows the GT segmentation.}
\label{pan_res_vis}
\end{figure}

\begin{figure}[!t]
\centering
\begin{tabular}{cc}
\thead{RGB} & \thead{Depth \\ Completion} \\
{\includegraphics[width = 1.8in]{depth_results/00050.jpg}} &
{\includegraphics[width = 1.8in]{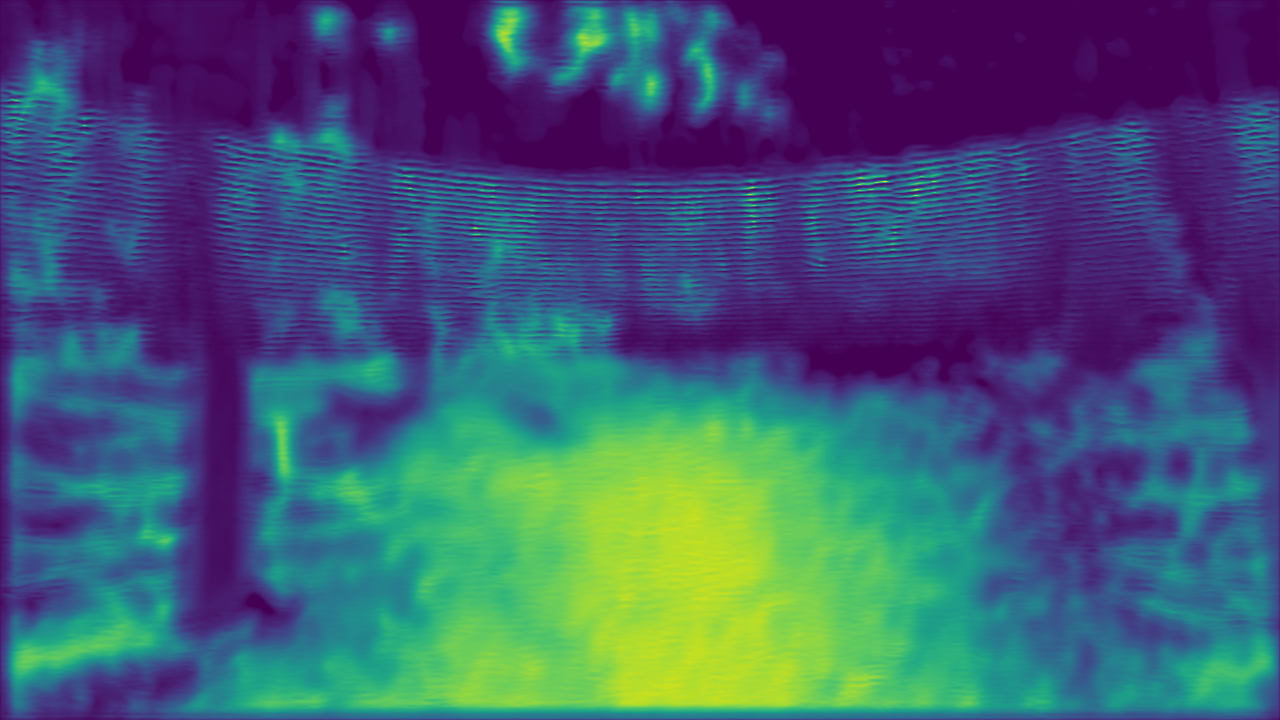}} \\
{\includegraphics[width = 1.8in]{depth_results/00070.jpg}} &
{\includegraphics[width = 1.8in]{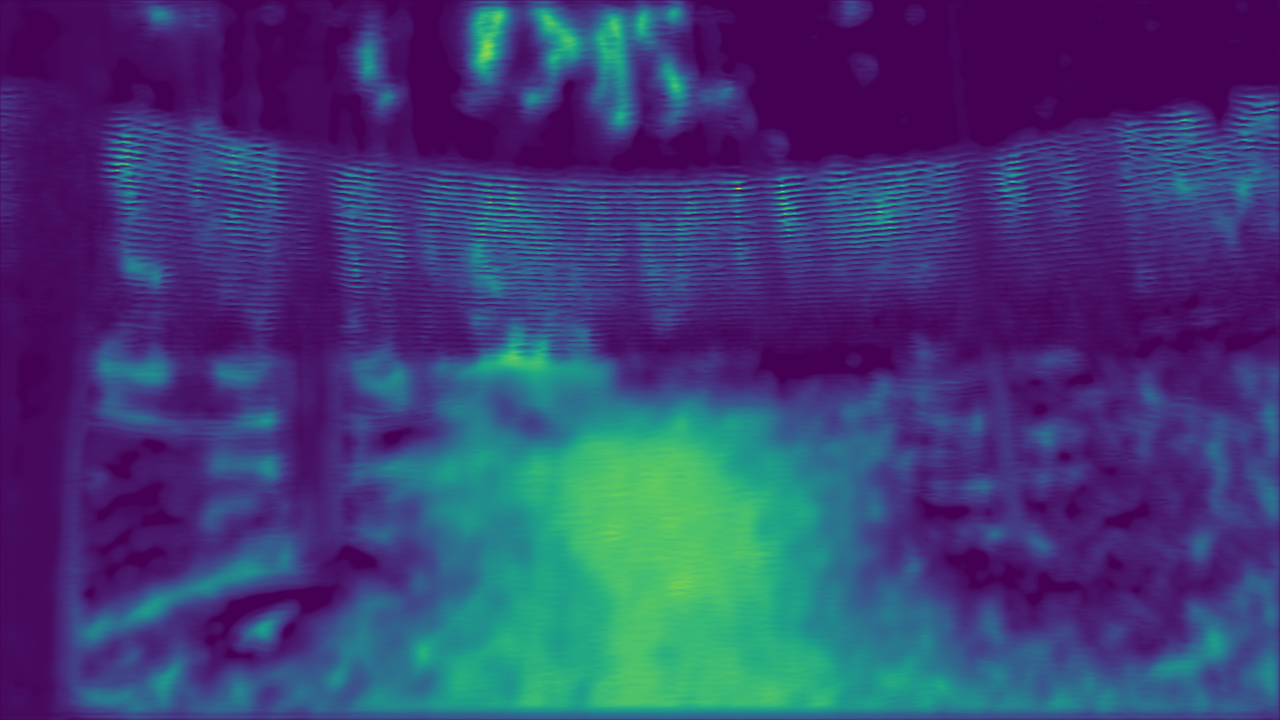}} \\
{\includegraphics[width = 1.8in]{depth_results/00099.jpg}} &
{\includegraphics[width = 1.8in]{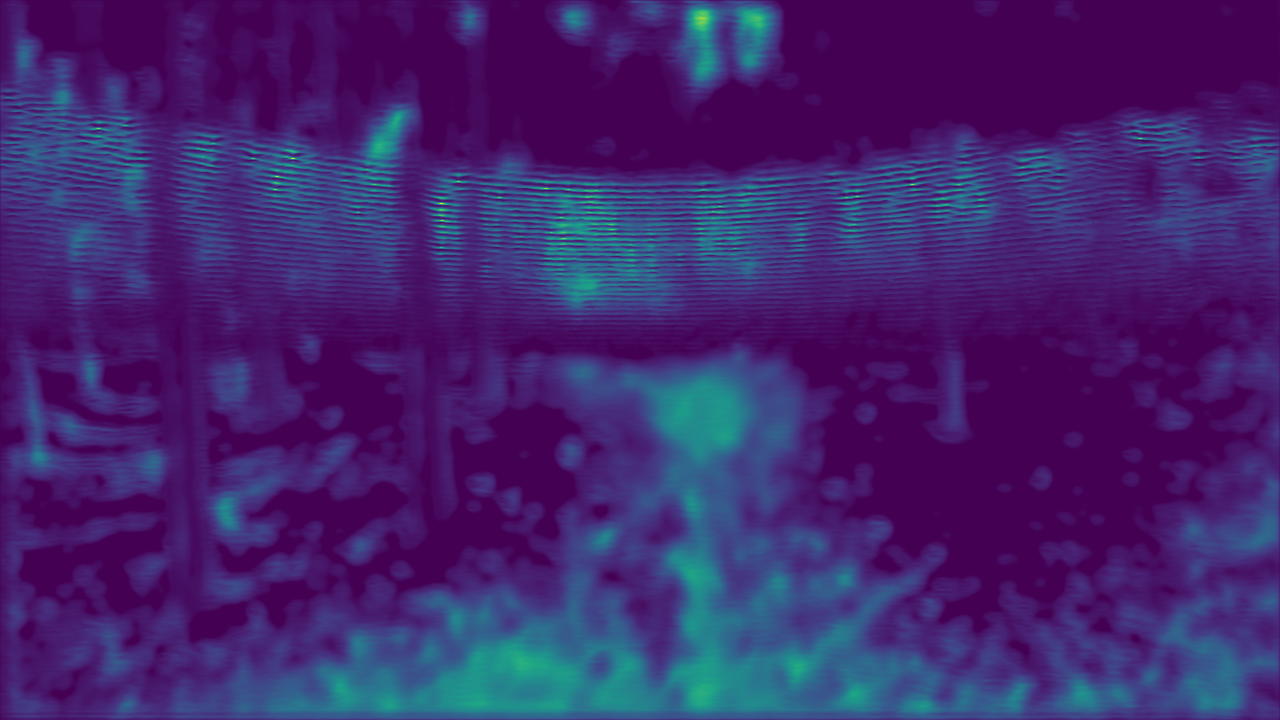}} \\
\end{tabular}
\caption{FuseNet \cite{2d3d} Qualitative Results. The first column shows three different input images and the second column depicts the corresponding depth completion results for every input scene. FuseNet \cite{2d3d} generalizes fairly well, including the areas where no depth information was provided on the sparse depth maps, however some of the fine edges and structures from the objects in the forest could not be recovered.}
\label{depth_res_vis}
\end{figure}

The quantitative performance results of every model are shown in Table \ref{quant_results}. The overall performance of Mask R-CNN \cite{maskrcnn} as measured by the mAP@50 is $28\%$, and from Figure \ref{mask_res_vis}, it is clear that, while it can detect tree trunks, the segmentation masks are not highly accurate, especially when the tree trunks are very close to each other. This reveals an opportunity for improvement in densely populated forests. Similarly, EfficientPS \cite{effps} detects tree trunks effectively, as shown in Figure \ref{pan_res_vis}. Nonetheless, segmentation masks are not tightly bound to the corresponding trees in dense forest scenes. The overall PQ reached $27.8\%$; however, there is a significant difference when the PQ is evaluated separately for "things" and "stuff". EfficientPS performance is poorer when evaluated on "things" instances, which holds for the SQ and RQ metrics as well. These results again highlight the challenges that dense forest scenarios pose to machine learning methods and data-driven algorithms, especially segmenting single objects. We also computed the mAP@50 for EfficientPS, and it is noticeable that it outperforms Mask R-CNN in the same metric. On the other hand, FuseNet \cite{2d3d} generalizes reasonably well, even over the areas within the images where no sparse depth information is provided. However, fine structures and boundaries of objects like trees are lost, as shown in Figure \ref{depth_res_vis}; hence there is also room for improvement in depth completion.

\begin{table}[h]
\centering
\caption{Quantitative Results}\label{quant_results}
 \begin{tabular}{ ||c|c|c||c|| } 
\hline
Model & Metric & Value \\
\hline\hline
\multirow{1}{8em}{Mask R-CNN} & mAP@50 & $28\%$  \\ 
\hline
\multirow{10}{8em}{EfficientPS} & PQ & 27.8 \\ 
& SQ & 32.5\% \\ 
& RQ & 34.1\% \\ 
& PQ Stuff  & 41.5\% \\ 
& SQ Stuff & 41.5\% \\ 
& RQ Stuff & 50.0\% \\ 
& PQ Things & 18.7\% \\ 
& SQ Things &  26.5\%  \\ 
& RQ Things & 23.5\% \\ 
& mAP@50 & 50.0\% \\ 
\hline
\multirow{1}{8em}{FuseNet} & RMSE(mm) & 489.96 \\ 
\hline

\end{tabular}

\end{table}

\section{Conclusion}

In this paper, we introduced \textit{FinnWoodlands}, a unique dataset for scene understanding in forest environments. \textit{FinnWoodlands} provides manual GT annotations for instance segmentation, semantic segmentation, and panoptic segmentation in addition to sparse depth maps, which are necessary for holistic scene representation. Our dataset contains unstructured objects commonly found in forest scenarios and focuses on detecting and segmenting tree trunks from three different tree species, namely "Spruce", "Birch", and "Pine" tree trunks. We collected data with a relatively simple data-collection setup which can easily be replicated to produce similar data and extend \textit{FinnWoodlands} dataset. We also provided an initial benchmark by testing our data with three deep neural network architectures, Mask R-CNN \cite{maskrcnn} for instance segmentation, EfficientPS \cite{effps} for panoptic segmentation, and FuseNet \cite{2d3d} for depth completion. The results reveal some of the challenges of computer vision when models are deployed in very unstructured scenarios such as forests, thus highlighting opportunities for improvement in similar scenarios. Our dataset can potentially impact the development of forestry applications and research in the field of computer vision in forest-like scenarios.

%
%
%
\bibliographystyle{splncs04}
\bibliography{mybibliography}
%




\end{document}